\pdfoutput=1
\documentclass[11pt]{article}

\usepackage[final]{acl}

\usepackage{times}
\usepackage{latexsym}

\usepackage[T1]{fontenc}

\usepackage[utf8]{inputenc}

\usepackage{microtype}

\usepackage{inconsolata}

\usepackage{graphicx}

\usepackage{comment}
\usepackage{bm}
\usepackage{enumitem}
\usepackage{amsmath}        % underset
\usepackage{amssymb}

\usepackage{mathtools}
\usepackage{wrapfig, lipsum}
\usepackage{adjustbox}

\usepackage{graphicx}
\usepackage{multirow, booktabs}
\usepackage{makecell}
\usepackage{subcaption}
\usepackage{bbm}

\usepackage{scalerel} %for scaling (e.g. stretching)
\usepackage{tikz}

\DeclareMathOperator*{\argmax}{arg\,max}

\newcommand{\transpose}{{\scriptstyle\sf T}}
\newcommand{\std}[1]{\scriptsize{$\pm$#1}}
\newcommand{\smallcaption}{\fontsize{8.3}{9}\selectfont}
\newcommand{\n}{\textbackslash n}
\newcommand{\shortminus}{\scalerel*{-}{0.2}}
\newcommand{\shortdash}{\hspace{1pt}\tikz[baseline=-0.5ex]\draw[line width=0.4pt] (0,0) -- (0.17,0);\hspace{0.5pt}} % Adjust the length (0.2) as needed
\newcommand{\shortplus}{\hspace{1pt}\tikz[baseline=-0.5ex]\draw[line width=0.4pt] (0,-0.1) -- (0,0.1) (0,-0.1) -- (0,0.1) (0.1,0) -- (-0.1,0);\hspace{0.5pt}}

\newcommand{\shortequal}{\hspace{1pt}\tikz[baseline=-0.5ex]\draw[line width=0.4pt] (0,0.04) -- (0.17,0.04) (0,-0.04) -- (0.17,-0.04);\hspace{0.5pt}}

\newcommand{\vshortdash}{\hspace{1pt}\tikz[baseline=-0.5ex]\draw[line width=0.3pt] (0,0) -- (0.13,0);\hspace{0.5pt}} % Adjust the length (0.2) as needed

\newcommand{\smallequal}{\resizebox{0.7\width}{\height}{$=$}}

\title{Efficient LLM Comparative Assessment: \\A Product of Experts Framework for Pairwise Comparisons}

\author{
    Adian Liusie \\
    University of Cambridge \\
    \texttt{al826@cam.ac.uk} \\
    \And
    Vatsal Raina \\
    University of Cambridge \\
    \texttt{vr311@cam.ac.uk} \\
    \AND
    Yassir Fathullah \\
    University of Cambridge\\
    \texttt{yf286@cam.ac.uk} \\
    \And
    Mark  J. F. Gales \\
    University of Cambridge\\
    \texttt{mjfg@eng.cam.ac.uk} \\
}

\begin{document}
\maketitle
\begin{abstract}
    LLM-as-a-judge approaches are a practical and effective way of assessing a range of text tasks. However, when using pairwise comparisons to rank a set of candidates, the computational cost scales quadratically with the number of candidates, which has practical limitations. This paper introduces a Product of Expert (PoE) framework for efficient LLM Comparative Assessment. Here individual comparisons are considered experts that provide information on a pair's score difference. The PoE framework combines the information from these experts to yield an expression that can be maximized with respect to the underlying set of candidates, and is highly flexible where any form of expert can be assumed. When Gaussian experts are used one can derive simple closed-form solutions for the optimal candidate ranking, and expressions for selecting which comparisons should be made to maximize the probability of this ranking. Our approach enables efficient comparative assessment, where by using only a small subset of the possible comparisons, one can generate score predictions that correlate well with human judgements. We evaluate the approach on multiple NLG tasks and demonstrate that our framework can yield considerable computational savings when performing pairwise comparative assessment. With many candidate texts, using as few as 2\% of comparisons the PoE solution can achieve similar performance to when all comparisons are used.\footnote{code available at: \url{https://github.com/adianliusie/PoE-LLM-comparative-assessment}}
\end{abstract}

\section{Introduction}
The advent of instruction-following \cite{wei2021finetuned, ouyang2022training} Large Language Models (LLMs) \cite{brown2020language, touvron2023llama} has enabled systems to exhibit impressive zero-shot capabilities on a range of Natural Language Processing (NLP) tasks. One such practical application is in Natural Language Generation (NLG) evaluation \cite{fabbri2021summeval}, where LLMs can be prompted to assess the quality of texts for particular attributes \cite{wang2023chatgpt, liu-etal-2023-g, zheng2023judging}. A popular approach is LLM comparative assessment, where pairwise comparisons are used to determine which of two texts is better \cite{zheng2023judging, qin2023large, liusie-etal-2024-llm}. Although using pairwise comparisons has been shown to better align with human preferences \cite{liusie-etal-2024-llm} than LLM scoring approaches \cite{wang2023chatgpt, liu-etal-2023-g}, the set of all comparisons scales quadratically with the number of inputs, which may be impractical in real-world use cases. Therefore, one may instead consider methods that only use a subset of comparisons to predict the scores, such that performance is maintained in computationally efficient settings.

Due to its applicability to sports, search and many other domains, the task of going from a subset of comparisons to a final ranking/scoring has been well-studied and extensively explored \cite{davidson1976bibliography, david1963method, luce2005individual, cattelan2012models}. However, in the majority of set-ups, the comparative decisions are binary (win/loss, although occasionally also win/loss/tie). LLMs, however, not only provide the outcome of the comparison but also additional information, such as the associated probability that A is better than B. Despite this available information, current LLM comparative works often leverage naive metrics such as win-ratio \cite{qin2023large, zheng2023judging, liusie-etal-2024-llm} and average probability \cite{park2024paireval, molenda2024waterjudge}, with little analysis on how to maximally extract the information from the comparisons.

This paper introduces a theoretical framework for viewing comparative assessment that enables practical scoring even in cases when the full set of comparisons is not used. We conceptualize the process as a Product of Experts (PoE) \cite{hinton1999products, Welling:2007}, where each comparative decision is assumed to provide information on the quality difference between the two competing texts. The framework is highly flexible and can use any form of expert. By considering two forms of experts, namely 1) the Gaussian distribution with linear assumptions and 2) an extension of the Bradley-Terry (BT) model for soft probabilities (motivated by looking at its limiting behaviour), we demonstrate that the PoE framework for comparative assessment can achieve efficient and effective NLG assessment. With the Gaussian expert, the framework yields a closed-form solution for the scores, which conveniently yields standard metrics when using the full set of comparisons. We demonstrate that our Product of Expert framework leads to significant performance boosts across models, datasets and assessment attributes, and even when using a fraction of the possible comparisons, can achieve high performance with minimal performance degradation from the full set. 

This paper makes several contributions. 1) We introduce the PoE perspective of comparative assessment, a highly flexible theoretical framework which enables one to directly model the distribution of scores given a set of comparisons. 2) We propose two experts, a soft Bradley-Terry expert (by considering the limiting behaviour of BT) and a Gaussian expert that has closed-form solutions and can be used to select the most informative comparisons. 3) We demonstrate practically that the PoE solution yields significant computational savings and empirically show that convergence is reached significantly faster than when using other baseline approaches for several datasets.

\section{Background and Related Work}
\textbf{Traditional/Tailored NLG Evaluation}: Initially, the outputs of NLG systems were evaluated against ground-truth human-annotated references, using N-gram overlap metrics \cite{papineni2002bleu, lin2004rouge, banerjee2005meteor} or similarity metrics \cite{zhang2019bertscore}. For more fine-grained evaluation, later studies developed bespoke evaluators for particular task dimensions such as summary consistency \cite{wang2020asking, manakul2023mqag, kryscinski2020evaluating} or dialogue coherence \cite{dziri2019evaluating, ye2021towards}. Further extensions considered unified evaluators, which evaluate multiple independent attributes \cite{mehri2020usr, yuan2021bartscore, zhong2022towards}. A drawback with these traditional NLG evaluation approaches is that they typically are bespoke towards particular tasks and attributes and, therefore, cannot easily be extended to new domains.  \vspace{1mm}

\noindent\textbf{LLM-Based NLG Evaluation}: Given the impressive instruction-following \cite{ouyang2022training, chung2022scaling} capabilities of LLMs such as GPT-4 \cite{achiam2023gpt} and open-sourced variants \cite{chung2022scaling, touvron2023llama}, recent works have studied leveraging these LLMs for general zero-shot NLG evaluation. Methods include GPTScore \cite{fu2023gptscore}, which computes the LLM likelihood of generating the response, and LLM-as-a-judge approaches \cite{zheng2023judging} that prompt models to provide scores \cite{wang2023chatgpt, kocmi2023large, liu-etal-2023-g} or use pairwise comparisons to determine which of two responses is better \cite{qin2023large, liusie-etal-2024-llm}. %These methods work effectively for a range of tasks and datasets and also yield performance that is competitive, if not better, than traditional bespoke solutions \cite{liusie-etal-2024-llm}. \\
\vspace{1mm}

\noindent\textbf{LLM Comparative Assessment}: Various recent works have used pairwise LLM comparative assessment for ranking texts: \citet{liusie-etal-2024-llm} demonstrate that for moderate-sized LLMs, comparative assessment outperforms LLM scoring as well as various bespoke baselines. They compute the win-ratio using all $N(N\!-\!1)$ comparisons as well as with a subset of comparisons (where large degradations are observed). Further, \citet{qin2023large} use pairwise comparisons for retrieving relevant sources, both using the full set of comparisons as well as sorting-based algorithms. \citet{park2024paireval} apply comparative assessment to dialogue evaluation, computing the average probability over a randomly sampled set of comparisons as the score quality. They also adapt the model with supervised training. Lastly, \citet{liu2024aligning} demonstrate limitations for LLM scoring and, therefore, instead consider pairwise comparisons. They introduce PAirwise-preference Search (PAIRS), a variant of the merge sort algorithm using LLM probabilities. \vspace{1mm}

\noindent\textbf{Comparisons to Scores}: Although LLMs have only recently been used as pairwise evaluators, the problem of ranking a set of candidates from a set of pairwise comparisons has been extensively studied in many different contexts, including sports \cite{beaudoin2018computationally, csato2013ranking}, information retrieval \cite{cao2007learning, liu2009learning} and social studies \cite{manski1977structure, louviere2000stated}. Arguably the most widely used parametric model is the Bradley-Terry model \cite{bradley1952rank}, which models the win probabilities based on the difference of the latent scores of the compared items. The latent scores are deduced by maximizing the likelihood of the observed pairwise comparison data, with various works discussing algorithms that converge to the solution \cite{davidson1976bibliography, david1963method, cattelan2012models}. Additionally, \cite{chen2022optimal} investigate predicting rankings under the Bradley-Terry-Luce model \cite{luce2005individual}, while
TrueSkill \cite{herbrich2006trueskill, minka2018trueskill} extends the Bradley-Terry model to incorporate uncertainties in player skills (in a sports context) under a Bayesian framework.

\section{A Product of Experts Perspective of Comparative Assessment}
\label{sec:poe}

%\subsection{Task Definition}
Let $x_{1:N} \in \mathcal{X}$ be a set of $N$ candidate texts and $s_{1:N} \in \mathbb{R}$ the scores of the texts for a particular assessed attribute. Given a set of $K$ pairwise comparisons, $\mathcal{C}_{1:K}$, the objective is to determine a predicted set of scores, $\hat{s}_{1:N}$, that are close to the true scores, $s^*_{1:N}$.

% Let $x_{1:N} \!\in\! \mathcal{X}$ be a set of $N$ candidate texts and $s_{1:N} \!\in\! \mathbb{R}$ the respective texts' scores for an assessed attribute. Given a set of $K$ pairwise comparisons, $\mathcal{C}_{1:K}$, the objective is to determine a predicted set of scores, $\hat{s}_{1:N}$, close to the true scores, $s^*_{1:N}$.

\subsection{The Bradley–Terry Model} For traditional comparative assessment set-ups, outcomes are usually discrete and either binary (win/loss) or ternary (win/draw/loss). A standard approach of going from a set of discrete comparisons $\mathcal{C}_{1:K}$ to predicted scores $\hat{s}_{1:N}$ is the Bradley–Terry model \cite{bradley1952rank, zermelo1929berechnung}. Assuming each comparison $C_k$ is of the form $(i, j, y_{ij})$, where $y_{ij}\in \{0, 1\}$ represents a draw from a binomial distribution which depends on the "quality" of the two texts. Here the probability that the quality of $x_i$, $z_i$, is deemed to be better than the quality of $x_j$, $z_j$, can be expressed as  ${\tt P}(z_i\!\succ\! z_j| s_i \!-\! s_j) = \sigma(s_i \!-\! s_j)$. The most popular form is the sigmoid function, $\sigma(x) = 1/(1+e^{-x})$. 
The Bradley-Terry model treats the scores as parameters of the model, and aims to maximize the likelihood of the binomial draws,

% \begin{gather}
%     {\tt P}(\mathcal{C}_{1:K}| s_{1:N})=\!\!\! \prod_{i, j \in \mathcal{C}_{1:K}}  \!\!\! {\tt P}(y_{ij}|s_{1:N}) \\ 
%     {\tt P}(y_{ij}|s_{1:N}) \!=\! \sigma(s_i \shortdash s_j)^{y_{ij}} (1\shortdash \sigma(s_i \shortdash s_j))^{1\shortdash y_{ij}} 
%     \label{eq:bradley_terry} 
% \end{gather}
% \vspace{-7mm}
% \begin{equation}
%         \hat{s}_{1:N} = \argmax_{s_{1:N}} {\tt P}(\mathcal{C}_{1:K}| s_{1:N}) 
% \end{equation}
%
\begin{gather}
    \hat{s}_{1:N} = \argmax_{s_{1:N}} \prod_{i, j \in \mathcal{C}_{1:K}}  \!\!\! {\tt P}(y_{ij}|s_{1:N}) 
    \label{eq:bradley_terry} 
\end{gather}
%%where,
\begin{gather}
    {\tt P}(y_{ij}|s_{1:N}) \!=\! \sigma(s_i \shortdash s_j)^{y_{ij}} (1\shortdash \sigma(s_i \shortdash s_j))^{1\shortdash y_{ij}} 
    \label{eq:bradley_terry} 
\end{gather}
\noindent Although no closed-form solution exists, Zermello's algorithm \cite{zermelo1929berechnung} can be used to iterate the solution until convergence is reached. Furthermore, while Zermello's algorithm is known to be slow to converge \cite{dykstra1956note, hunter2004mm}, later improvements have demonstrated faster convergence rates \cite{newman2023efficient}.

\subsection{A Product of Experts Perspective} 
\label{ssec:soft_bradley_terry}
For LLM comparative assessment, as opposed to traditional binary comparative decisions, one has access to richer information, including the associated probability of a decision. Each comparison outcome can therefore be extended to the form $(i, j, p_{ij})$ where $p_{ij} \!=\! {\tt P}_{\tt lm}(z_i\!\succ\! z_j| x_i, x_j)$, the LLM probability of the comparative decision. To conveniently incorporate the soft-probability observations, we explore directly modelling the probability of scores given the comparative observations and reformulate the scores as a Product of Experts. A Product of Experts (PoE) \cite{hinton1999products, Welling:2007} combines the information gained from many individual experts by taking their product and normalizing the result. One can consider each comparison as information gained from independent experts, enabling the probability for the scores to be written as:
\begin{equation}
    {\tt p}(s_{1:N} | \mathcal{C}_{1:K}) = \frac{1}{Z} \prod_{i, j \in \mathcal{C}_{1:K}} {\tt p}(s_i \! - \! s_j | C_k) 
\end{equation}
Each expert can be conditioned on the observed LLM probability such that ${\tt p}(s_i \! - \! s_j | C_k) = {\tt p}(s_i \! - \! s_j | p_{ij})$. As a possible expert, we consider a form related to the limiting behaviour of the Bradley-Terry Model and re-express Equation \ref{eq:bradley_terry} with a probabilistic classification result form, 
\begin{align}
   {\tt p}(s_i \shortdash s_j| p_{ij}) \!&=\!\! \frac{1}{Z_{ij}} \sigma(s_i \shortdash s_j)^{p_{ij}} (1\shortdash \sigma(s_i \shortdash s_j))^{1\shortdash p_{ij}} \notag
   %& = \frac{1}{Z_{ij}} \cdot \frac{e^{p(s_i - s_j)}}{1 + e^{(s_i - s_j)}}
   \label{eq:bradley_terry_soft}
\end{align}
Defined within the range $0\!<\!p_{ij}\!<\!1$, where $Z_{ij}=\pi \!/\! \sin(p_{ij} \pi)$ is a normalization term to ensure a valid probability density function. The solution can similarly be found using Zermelo's algorithm. Although the resulting expression is difficult to analyze, one can apply a Laplace approximation to approximate the score distribution as a Gaussian (shown in Appendix \ref{sec:laplace_approx}), which yields a more intuitive expression that can be useful for downstream applications.

\subsection{Properties of Gaussian Experts} The experts are not restricted to sigmoid-based modelling, and one can select any family of probability distributions. One option is to directly model Gaussian Experts, which have convenient properties such as a closed-form expression for the PoE solution \cite{zen2011product}. If the underlying distribution is assumed to be Gaussian with the mean $f_{\mu}(p_{ij})$ and variance $f_{\sigma}(p_{ij})$ only dependent on the comparative probability, such that ${\tt p}(s_i \! - \! s_j | p_{ij}) = \mathcal{N}{\big (}s_i\!-\!s_j; f_{\mu}(p_{ij}), f_{\sigma}(p_{ij}){\big )}$, then by representing the scores in vector form, ${\bf s}\!=\![s_{1:N}]$, one can express the distribution as, 
\begin{equation}
    {\tt p} ({ \bf W} {\bf s}| \mathcal{C}_{1:K}) = \mathcal{N}\left({\bf W} {\bf s}; \bm{\mu}, \text{diag}(\bm{\sigma^2}) \right)
\end{equation}
Where ${\bf W} \!\in\! R^{K \times N}$ (illustrated in Appendix \ref{sec:W_example}) is a matrix representing the set of comparisons, such that for the $k^{\text{th}}$ comparison between $i$ and $j$ ${\bf W}_{ki} \!=\! 1$, ${\bf W}_{kj} \!=\! -1$, and ${\bf W}_{km} \!=\! 0 \; \forall m \! \neq \! i, j$, $\; {\bf s}$ is the N-dimensional column vector of $s_{1:N}$, $\bm{\mu} \!\in\! R^{K}$ is a vector of the means, and $\bm{\sigma^2} \!\in\! R^{K}$ equivalently represents the variances, 
\begin{align}
    \bm{\mu} &= [f_{\mu}(p_{ij}^{(1)}), f_{\mu}(p_{ij}^{(2)}), ... f_{\mu}(p_{ij}^{(K)})]^\transpose \\
    {\bm{\sigma}^2} &= [f_{\sigma}(p_{ij}^{(1)}), f_{\sigma}(p_{ij}^{(2)}), ... f_{\sigma}(p_{ij}^{(K)})]^\transpose
\end{align}
Note that as defined, any shift of the scores $\bm{s}$ will yield an equivalent output. To address this, an additional expert on the first element can be added, such that ${\tt p}(s_1|\mathcal{C}_{0}) = \mathcal{N}(0, \sigma_0^2)$,
%
% \begin{equation}
%     {\tt p}(s_1|\mathcal{C}_{0}) = \mathcal{N}(0, \sigma_0^2) 
% \end{equation} 
%
prepending an extra row to all of ${\bf W}$, $\bm{\mu}$ and $\bm{\sigma^2}$, yielding ${\bf \tilde{W}}$, $\bm{\tilde{\mu}}$ and $\bm{\tilde{\sigma}^2}$ respectively. The distribution takes a similar form, ${\tt p}({ \bf \tilde{W}} {\bf s}| \mathcal{C}_{1:K}) = \mathcal{N}({\bf \tilde{W}} {\bf s}; \bm{\tilde{\mu}}, \text{diag}(\bm{\tilde{\sigma}^2}))$, which can be rearranged to yield a Gaussian expression for the score distribution, ${\tt p}({s}_{1:N}| C_{1:K}) = \mathcal{N}({\bf s}; {\bm{\mu}_s^*}, {\bf \tilde \Sigma}^*_s)$, with mean and covariance matrix defined as,
\begin{align}
    {\bm{\mu}_s^*} &= {\bf \tilde{W}}^\transpose {\bf \tilde \Sigma}^{-1} {\bf \tilde{W}})^{-1} {\bf \tilde{W}}^\transpose {\bf \tilde \Sigma}^{-1} \bm{\tilde{\mu}} \\
    {\bf \tilde \Sigma}^*_s &= ({\bf \tilde{W}}^\transpose {\bf \tilde \Sigma}^{-1} {\bf \tilde{W}})^{-1}
\end{align}
% \begin{align}
%     {\tt p}({s}_{1:N}| C_{1:K}) = \mathcal{N}\left({\bf s}; ({\bf \tilde{W}}^\transpose {\bf \tilde \Sigma}^{-1} {\bf \tilde{W}})^{-1} {\bf \tilde{W}} {\bf \tilde \Sigma}^{-1} \bm{\tilde{\mu}} \; , \; ({\bf \tilde{W}}^\transpose {\bf \tilde \Sigma}^{-1} {\bf \tilde{W}})^{-1} \right)
% \end{align}
%
where ${\bf \tilde \Sigma} = \text{diag}(\bm{\tilde{\sigma}^2})$  (the rearranging is shown in Appendix \ref{sec:rearranging_gaussian}). Therefore, the mean of the Gaussian provides a simple and closed-form solution to the maximum probability solution, $\hat{s}_{1:N}$, 
\begin{align}
    {\bf \hat{s}} &= \argmax_{{s}_{1:N}} {\tt p}({s}_{1:N}|\mathcal{C}_{1:K}) \\
    &= ({\bf \tilde{W}}^\transpose {\bf \tilde \Sigma}^{-1} {\bf \tilde{W}})^{-1} {\bf \tilde{W}}^\transpose {\bf \tilde \Sigma}^{-1} \bm{\tilde{\mu}}
\end{align}
%
%\vspace{1mm}
\subsection{Further Gaussian Assumptions}
\label{ssec:gaussian_assumptions}

A drawback with the Gaussian Expert is that producing $\bm{\tilde{\mu}}$ and $\bm{\tilde{\sigma}^2}$ requires knowledge of both $f_{\mu}(p)$ and $f_{\sigma}(p)$. This is not available without human-annotated data, making the approach impractical for zero-shot applications. To enable a practical solution applicable in zero-shot settings, one can make two assumptions on the Gaussian experts: 1) that the variance is constant regardless of the predicted probability $f_{\sigma}(p) = \sigma^2$, and 2) that the mean scales linearly with the probability $f_{\mu}(p) = \alpha \! \cdot \! (p - \beta)$. These assumptions appear reasonable for several models and datasets (in Appendix Figure \ref{fig:lienar_assumptions}) and simplify the solution to,
\begin{equation}
    {\bf \hat{s}} = \alpha \cdot ({\bf \tilde{W}}^{\transpose} {\bf \tilde{W}})^{-1} {\bf \tilde{W}}^\transpose  \bm{\tilde{\mu}}
    \label{eq:}
\end{equation} 
where $\bm{\tilde{\mu}}^\transpose \!=\! [0, p_{ij}^{(1)} \!\!-\! \beta, ..., p_{ij}^{(K)} \!\!-\! \beta]$. Note that a sensible choice might be $\beta\!=\!0.5$, since when inputting texts of equal quality into an unbiased system, an average output probability of 0.5 would be expected. Further, the value of $\alpha$ only influences the relative spacing and subjective scale used to score the texts and can arbitrarily be set to 1. 

\subsection{Modelling Bias in Non-Symmetric Settings}
LLMs can have inconsistent outputs where $p_{ij} \! \neq \! (1\!-\!p_{ji})$ and, in particular, demonstrate positional bias \cite{zheng2023judging, chen2024humans, liusie2024teacher}. Positional bias occurs when the system prefers one position over another such that $\mathbb{E}_{{\tt p}_{\text{lm}}(p)}[p] \!\neq\! 0.5$, while for unbiased systems, the expectation should be near 0.5. Combining the probabilities from both permutations such that $\tilde{p}_{ij} \!=\! \frac{1}{2}\!\cdot\!(p_{ij}\!+\!(1\!-\!p_{ji}))$ ensures that $\tilde{p}_{ij} \!=\! (1\!-\!\tilde{p}_{ij})$ and eliminates positional bias; however, it requires two LLM calls per comparison and may not be the best use of LLM calls. To efficiently minimize the impact of positional bias without requiring both LLM permutation calls, we investigate directly modelling model position bias into the experts. A simple approach is to introduce a bias parameter $\gamma$ that shifts the experts such that, ${\tt p}_{\gamma}(s_i - s_j | p_{ij}) = {\tt p}(s_i \!- \! s_j \! - \gamma | p_{ij})$. The value of $\gamma$ can be determined by noting that the expected score difference between two randomly sampled texts is zero, $\mathbb{E}[s_i - s_j]=0$. For the linear Gaussian expert, this is equivalent to applying a linear shift in the mean, and therefore by considering $\mathcal{N}{\big (}s_i\!-\!s_j; \alpha \! \cdot \! (p_{ij}-\beta), \sigma^2{\big )}$,
\begin{align}
    \mathbb{E}[s_i - s_j] = \mathbb{E}[f_{\mu}({p_{ij}})] = \alpha\big( \mathbb{E}[p_{ij}] - \beta \big)
\end{align}
% \begin{align}
%     \mathbb{E}_{i \neq j}[s_i - s_j] = \mathbb{E}_{i \neq j}[f_{\mu}({p_{ij}})]  = \mathbb{E}_{i \neq j}[\alpha(p_{ij}-\beta)] = \alpha\big( \mathbb{E}_{i\neq j}[p_{ij}] - \beta \big)
% \end{align}
setting the expression to zero yields that the debiasing term $\beta \!=\! \mathbb{E}[p_{ij}]$. For Bradley-Terry, though it can be shown that $f_{\mu}({p_{ij}})=- \pi \! \cdot \! \cot(\pi p_{ij})$, this value tends to infinity when $p_{ij}$ approaches either $0$ or $1$. Therefore, instead of setting the expected value of the skill difference for any random pair to be zero, we approximate finding the bias by ensuring the mode of the underlying (log-) distribution is 0 when the skill difference is 0. Based on this approximation, the resulting bias parameters for the extended Bradley-Terry is $\gamma = -{\tt logit}(\mathbb{E}[p_{ij}])$ (see Appendix \ref{ssec:accounting_for_bias_bt} for further details).

\subsection{Comparison Selection}
\label{ssec:optimal}
The previous theory detailed how to determine the predicted scores $\hat{s}_{1:N}$ given a random set of observed comparisons $\mathcal{C}_{1:K}$. As an extension, one may consider how to select the set of comparisons that provide the most information. Under the Gaussian model, the probability of the most likely set of scores is given as, \vspace{-5mm}
\begin{equation}
    {\tt p}({\hat{s}}_{1:N}|\mathcal{C}_{1:K}) = \frac{\sqrt{{\tt det} ({\bf \tilde{W}}^{\transpose} {\bf \tilde{W}}) }}{ (2 \pi \sigma^2)^{N/2} }
\end{equation}
shown in Appendix \ref{sec:rearranging_gaussian}. For a fixed number of comparisons $K$, one may therefore aim to find the matrix ${\bf \tilde{W}}^*$ that minimizes the uncertainty, \vspace{-2mm}
\begin{align}
    {\bf \tilde{W}}^* = \argmax_{{\bf \tilde{W}}} {\tt p}({\hat{s}}_{1:N}|\mathcal{C}_{1:K}) \\ 
    \equiv \argmax_{{\bf \tilde{W}}} {\tt det} ({\bf \tilde{W}}^{\transpose} {\bf \tilde{W}})
\end{align}
This can be approximated through an iterative greedy search. Assume that ${\bf \tilde{W}}^{(k)*}$ is the selected comparison matrix using $k$ comparisons and ${\bf A}^{(k)}\!=\!({\bf \tilde{W}}^{(k)*^\transpose} {\bf \tilde{W}}^{(k)*})^{-1}$. The next selected comparison $(\hat{i}, \hat{j})$ can be calculated as, \vspace{-2mm}
\begin{equation}
    \hat{i}, \hat{j} = \argmax_{i, j} {\bf A}^{(k)}_{ii} + {\bf A}^{(k)}_{jj} - 2 \cdot {\bf A}^{(k)}_{ij}  
\end{equation} 
Shown in Appendix \ref{sec:update_additional}, where it is also shown that the inverse matrix ${\bf A}^{(k+1)}$ can be updated efficiently from ${\bf A}^{(k)}$. Additionally, it was noted previously that the score distribution using soft Bradley-Terry experts can be approximated as a Gaussian using a Laplacian approximation. Doing so and then selecting greedy optimal decisions will yield a similar selection scheme, 
\begin{eqnarray}
   \lefteqn{\hat{i}, \hat{j} = \argmax_{i, j} {\big [}} \\
    &&
    \sigma(\hat{s}_i \shortdash \hat{s}_j) \! \cdot \! \sigma(\hat{s}_j \shortdash \hat{s}_i) \! \cdot \! \left(\!{\bf A}^{(k)}_{ii} \!+\! {\bf A}^{(k)}_{jj} \shortdash 2 \! \cdot \! {\bf A}^{(k)}_{ij}  \right) {\big ]}
    \nonumber
\end{eqnarray}
% \begin{equation}
%     \hat{i}, \hat{j} = \argmax_{i, j} \; \sigma(\hat{s}_i \shortminus \hat{s}_j) \cdot \sigma(\hat{s}_j \shortminus \hat{s}_i) \left({\bf A}^{(k)*}_{ii} + {\bf A}^{(k)*}_{jj} - 2 \cdot {\bf A}^{(k)*}_{ij}  \right)
% \end{equation}
%
As shown in Appendix \ref{sec:laplace_selection}, where $\hat{s}_{1:N}$ represent the current score predictions using the comparisons so far. 
Therefore, selecting comparisons under the Laplacian approximation of the soft Bradley-Terry model leads to a similar selection process but with an additional term of $\sigma(\hat{s}_i \shortminus \hat{s}_j) \cdot \sigma(\hat{s}_j \shortminus \hat{s}_i)$. This term implies that under this model, comparisons between texts of similar quality are expected to reveal the most information. However, this approach requires the solution to be computed at each step and does not have an efficient update formula. Therefore, running this selection mechanism may be significantly more computationally expensive than the analysis using linear Gaussian experts, and practically, the utility of this selection scheme will depend on the tradeoff between this computational expense and the cost of an LLM forward pass.

\section{Experimental Setup}

\subsection{Datasets}
We consider a range of NLG evaluation datasets which have available ground-truth scores. For summary evaluation we use \textbf{SummEval} \cite{fabbri2021summeval} which has 100 articles each with 16 machine-generated summaries evaluated on coherency ({\tt COH}), consistency ({\tt CON}), fluency ({\tt FLU}), and relevancy ({\tt REL}). For dialogue response generation, we use \textbf{TopicalChat} \cite{mehri2020usr} which has 60 dialogue contexts with six responses per context assessed on coherency ({\tt COH}), continuity ({\tt CNT}), engagingness ({\tt ENG}), and naturalness ({\tt NAT}). For question difficulty ranking, we use \textbf{CMCQRD} \cite{mullooly2023cambridge}, which has 658 multiple-choice reading comprehension questions annotated on question difficulty. Lastly, for story evaluation, we use \textbf{HANNA} \cite{chhun2022human} which has 1056 machine-generated stories annotated by humans on coherency ({\tt COH}), complexity ({\tt CMP}) and surprisingness ({\tt SUR}). For CMCQRD and HANNA we compare the texts across all 658/1056 texts.

\subsection{Methodology}
\label{ssec:methodology}
% focusing on PCC as we care about relative spacing
% will do both context level and absolute evaluation. Make it clear what we mean by absolute as not standard
% 

\textbf{Base Large Language Models} Three different families of opensourced LLMs are used as judge LLMs: FlanT5 (3B, 11B) \cite{chung2022scaling}, instruction-tuned Mistral (7B) \cite{jiang2023mistral} and Llama2-chat (7B, 13B) \cite{touvron2023llama}. \vspace{2mm}

\noindent\textbf{LLM Pairwise Probability Calculations} To get comparative probabilities, we follow \citet{liusie-etal-2024-llm} and use ${\tt P(A)}/({\tt P(A)}\!+\!{\tt P(B)})$. The symmetric set-up (where both permutations are done) is used unless stated otherwise, though in Section \ref{ssec:unbalanced} the non-symmetric set-up is investigated. \vspace{2mm}

%In the Appendix, we also investigate the PoE framework for imbalanced scenarios.

\noindent\textbf{Comparison Selection} When considering comparative assessment with a subset of comparisons, the base experiments use a randomly drawn set of comparisons such that each comparison is equally likely to be chosen. For a set of inputs $x_{1:N}$, we randomly select $K$ unique pairs $(x_i, x_j)$ to be judged by the LLM, ensuring that each text $x_i$ is involved in at least one comparison. Experiments begin with $K\!=\!2N$ comparisons and $K$ is incremented to the full set of comparisons, $K\!=\!N\!\cdot\!(N\!-\!1)$. \vspace{2mm}

\noindent\textbf{Scoring Methods} Several different methods of mapping a set of comparisons to scores are used in this paper, categorized into binary decision-based or probability-based. For binary decision methods, our first baseline is the \textbf{win-ratio} which calculates the number of comparisons won as the quality score, as used in \citet{qin2023large, liusie-etal-2024-llm, raina2024question}. The second baseline is the Bradley-Terry model, \textbf{BT}, \cite{bradley1952rank}, where the solution is found by Zermelo \cite{zermelo1929berechnung} with a convergence threshold of $1e^{-4}$. Since any candidate that wins/loses all games will have an infinite score, a prior of $1/(N\!-\!1)$ wins is added to each selected comparison. For the methods that leverage the LLM probabilities, the baseline is the average probability \textbf{avg-prob} of a text in all its comparisons, as used in \citet{park2024paireval, molenda2024waterjudge}. To better leverage the probabilistic information, our paper proposes to decompose the probability into a product of experts. We propose two variants; 1) \textbf{PoE-BT} which uses a variant of the Bradley-Terry model extended to soft probabilities (described in Section \ref{ssec:soft_bradley_terry}), and 2) \textbf{PoE-g} which uses the Gaussian expert with the linear mean and constant variance assumptions (described in Section \ref{ssec:gaussian_assumptions}). Lastly, the final method is \textbf{PoE-g-hard}, which applies the PoE-gaussian framework, however, using hard binary decisions and not the soft probabilities. \vspace{2mm}

\noindent\textbf{Evaluation} For SummEval and TopicalChat, the summary-level Spearman score is used as the assessment metric. For each context, we do pairwise comparisons using the LLM on the full set of $N(N\!-\!1)$ comparisons. We then simulate using a subset of comparisons by randomly selecting $K$ of these outcomes. This process is repeated 100 times for a particular number of total comparisons, $K$, and we calculate both the mean and standard deviation of performance over the entire dataset. For Hanna and CMCQRD, there is no context dependence and therefore the number of candidate texts is much larger, with $N\!=\!1050$ and $N\!=\!550$ respectively. As such as we sample 200,000 comparisons (all symmetric), which is only a subset of the total possible comparisons, and provide analysis by simulating randomly sampling further subsets of these comparisons. For each $K$, we run 20 indpendent runs and average performance. For both datasets, equivalent tables for Pearson are provided in Appendix \ref{sec:extra_results}.

\section{Results}
\subsection{SummEval and TopicalChat}

In this Section, we investigate whether the Product of Experts framework can yield performance boosts for SummEval and TopicalChat in efficient settings. SummEval has 16 candidates per context $(N\!=\!16)$ and therefore considering all possible comparisons takes 240 comparisons, which though feasible, can be quite costly. Table \ref{tab:summary_table} presents SummEval performance when only a subset of the comparisons are made, with the average Spearman rank correlation coefficient (SCC) over all contexts and attributes presented for different base LLMs. Equivalent tables for TopicalChat are provided in Appendix \ref{ssec:extra_results_table} where similar trends are seen. The following observations can be made:

\begin{table}[t]
    \centering
    \renewcommand\tabcolsep{2pt} % Adjust column spacing as needed
    \fontsize{10}{10}\selectfont % Set the font size and line spacing for table contents
    \begin{tabular}{l c c c c c c}
        \toprule
               &     & \multicolumn{2}{c}{Decisions}  & \multicolumn{3}{c}{Probabilities} \\
        \cmidrule(lr){3-4} \cmidrule(lr){5-7}
        System & $K$ & Win-r & BT & Avg-pr & PoE-BT & PoE-g \\
        \midrule
        \multirow{2}{*}{Llama2-7B} 
        & 48  & 21.6 & 23.4 & 24.0 & 26.8 & 26.6 \\
        & 240 & 27.8 & 27.9 & 28.4 & 28.4 & 28.4 \\
        \midrule
        \multirow{2}{*}{Llama2-13B} 
        & 48  & 30.8 & 33.1 & 33.7 & 37.7 & 37.3 \\
        & 240 & 39.3 & 39.3 & 39.3 & 39.3 & 39.3 \\
        \midrule
        \multirow{2}{*}{Mistral-7B} 
        & 48  & 29.7 & 31.9 & 31.1 & 33.2 & 32.8 \\
        & 240 & 38.1 & 38.1 & 37.7 & 37.7 & 37.7 \\
        \midrule
        \multirow{2}{*}{FlanT5-3B} 
        & 48  & 34.1 & 36.6 & 38.4 & 42.6 & 42.4 \\
        & 240 & 43.6 & 43.6 & 44.3 & 44.3 & 44.3 \\
        \midrule
        \multirow{2}{*}{FlanT5-11B} 
        & 48  & 31.2 & 33.4 & 34.7 & 38.5 & 38.4 \\
        & 240 & 40.0 & 40.0 & 40.5 & 40.5 & 40.5 \\
        \bottomrule
    \end{tabular}
    \caption{Spearman Correlations for SummEval, averaged over all attributes ({\tt COH}, {\tt CON}, {\tt FLU}, {\tt REL}). $K$ is the number of comparisons made, where $K\!=\!240$ is the full set of comparisons.}
    \label{tab:summary_table}
    \vspace{-5mm}
\end{table}

\begin{figure*}[h]
    \centering
    \begin{subfigure}[t]{0.33\textwidth}
        \centering
         \includegraphics[width=\textwidth]{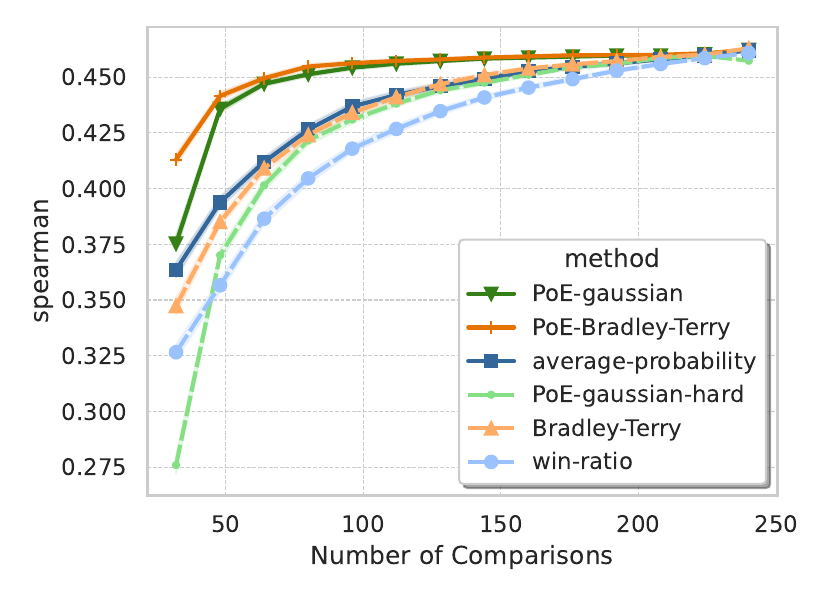}
        \caption{\smallcaption Llama2-13B, SummEval \texttt{REL}.}
        \label{fig:efficiency_curve_1}
    \end{subfigure}%
    ~ 
    \begin{subfigure}[t]{0.33\textwidth}
        \centering
         \includegraphics[width=\textwidth]{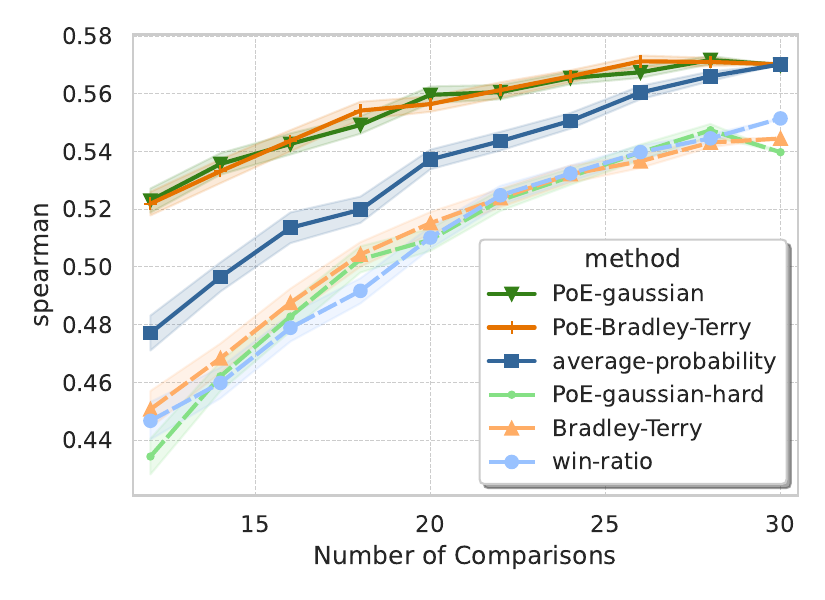}
        \caption{\smallcaption FlanT5-11B, TopicalChat, \texttt{ENG}.}
        \label{fig:efficiency_curve_2}
    \end{subfigure}%
    ~ 
    \begin{subfigure}[t]{0.33\textwidth}
        \centering
        \includegraphics[width=\textwidth]{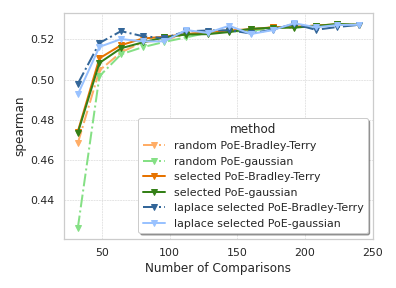}
        \captionsetup{format=plain, justification=centering}
        \caption{\smallcaption FlanT5-3B, SummEval \texttt{COH}, \\ \textbf{selected vs random}}
        \label{fig:selected_vs_random}
    \end{subfigure}
    \\
    \vspace{-3mm}
    \caption{Efficiency curves when sweeping $K$, the number of comparisons per context, where at each $K$ the comparisons are randomly drawn 100 times. Average performance with 95\% confidence is displayed.} 
    \vspace{-3mm}
\end{figure*}

\textbf{Average probability performs better than the win-ratio in efficient settings} When considering the full set of comparisons ($K\!=\!240$) the performance of average probability is only marginally better than using win-ratio (within 1 SCC). However, when using 20\% of the comparisons ($K\!=\!48$) the average probability yields significant gains of 3-4 SCC. This highlights that especially when only using a subset of comparisons, leveraging the soft probabilistic information is beneficial. 

\textbf{The PoE solution yields large gains in efficient settings} Even when only using hard decisions, for $K\!=\!48$, both the Bradley-Terry model (BT) and the PoE Gaussian with hard decisions (PoE-g-hard) have mild performance gains over the win-ratio. Nevertheless, the real benefits are seen when using PoEs with soft probabilities, with both PoE-BT and PoE-g significantly outperforming the average probability. With these methods, when using only 20\% of the comparisons, one can achieve performance close to when using the full comparison set (in four out of five cases within 2 SCC), when win-ratio would have degredations of up to 10 SCC. The findings are general and hold across the different SummEval attributes and models. 

\textbf{Gaussian PoE and BT PoE result in similar performing solutions} When using full-comparisons, the Gaussian PoE solution can be shown to be equivalent to the average probability (shown in Appendix \ref{ssec:poe-g-avg-prob-equivalence}) however the BT PoE approach will lead to a different solution. Nonetheless, the performance for both PoE-BT and PoE-g are very comparable for most models/datasets, in both the hard and soft set-ups. Further the Gaussian solution has the benefit of having a convenient closed form solution.

\textbf{Convergence rates} The results in Table \ref{tab:summary_table} showed performance for the arbitrary chosen operating point of $K\!=\!48$. Figures \ref{fig:efficiency_curve_1} and \ref{fig:efficiency_curve_2} show the performance for two models/attributes while sweeping $K$ from $K\!=\!N$ to the full set of comparisons, $K\!=\!N(N\!-\!1)/2$. The curves show that the performance improves smoothly while increasing number of comparisons, with the convergence rates considerably better with the PoE methods. Further plots for other models/tasks are provided in Appendix \ref{ssec:extra_efficient_plots_summeval}. 

% \begin{figure}[t]
%     \centering
%     \includegraphics[width=\columnwidth]{Figures/balanaced_summeval/llama2-13b-chat-summeval-relevance.pdf}
%     \vspace{-10mm}
%     \caption{\smallcaption Performance curve for Llama2-13B, SummEval \texttt{REL} when sweeping $K$ (with 95\% confidence intervals).}
%     \label{fig:efficiency_curves}
% \end{figure}

\subsection{Comparison Selection}
The previous results used random comparisons, however, an alternative would be to pre-select a set of comparisons that maximizes the information gained from a fixed number of comparisons. Section \ref{ssec:optimal} discusses how for the Gaussian-PoE, this can be achieved with a practical greedy approximation. Table \ref{tab:optimal_table} illustrates that at the operating point of $K\!=\!48$, pre-selecting the comparisons can provide further performance boosts, with the average performance of the probabilistic PoE approaches consistently increasing by $0.5$ SCC for all approaches, at no extra cost. Although the theory was derived using the Gaussian assumptions, the performance boosts are seen for all methods, with the largest gains for the win-ratio. Lastly, Figure \ref{fig:selected_vs_random} shows that performance gains are significant when few comparisons are made, but as the number of comparisons grows, the performance difference between random and optimal selection is negligible. Additionally, selecting the comparisons based on the Laplace-approximation of the Bradley Terry expert yields better performance when a small number of comparisons are considered; however, it is significantly more computationally expensive as the BT solution has to be determined at each timestep. 

\begin{table}[h]
    \centering
    \renewcommand\tabcolsep{2pt} % Adjust column spacing as needed
    \fontsize{9}{9}\selectfont % Set the font size and line spacing for table contents
    \begin{tabular}{l l cccc}
        \toprule
        System & Method & Win-r & Avg-pr & PoE-BT & PoE-g \\
        \midrule
        \multirow{2}{*}{Llama2-7B} 
        & Random  & 21.6 & 24.0 & 26.8 & 26.6 \\
        & Selected & 23.0 & 24.5 & 27.3 & 27.2 \\
        \midrule
        \multirow{2}{*}{Llama2-13B} 
        & Random  & 30.8 & 33.7 & 37.7 & 37.3 \\
        & Selected & 32.4 & 34.6 & 38.2 & 38.0 \\
        \midrule
        \multirow{2}{*}{Mistral-7B} 
        & Random  & 29.7 & 31.1 & 33.2 & 32.8 \\
        & Selected & 31.4 & 32.2 & 34.0 & 33.9 \\
        \midrule
        \multirow{2}{*}{FlanT5-3B} 
        & Random  & 34.1 & 38.4 & 42.7 & 42.4 \\
        & Selected & 36.0 & 39.3 & 43.2 & 42.9 \\
        \midrule
        \multirow{2}{*}{FlanT5-11B} 
        & Random  & 31.2 & 34.7 & 38.4 & 38.4 \\
        & Selected & 33.1 & 35.7 & 39.2 & 39.0 \\
        \bottomrule
    \end{tabular}
    \caption{SummEval Spearman correlations when using the greedy optimal set of comparisons, for $K\!=\!48$.}
    \label{tab:optimal_table}
    \vspace{-4mm}
\end{table}

% \begin{figure}[t]
%     \centering
%     \includegraphics[width=\columnwidth]{Figures/optimal_selection/selection_flant5-xl-summeval-coherency.pdf}
%     \vspace{-10mm}
%     \caption{\smallcaption FlanT5-3B, SummEval \texttt{COH}, selected vs random}
%     \label{fig:optimal_curve}
% \end{figure}

\subsection{Hanna and CMCQRD}

\begin{figure}[h]
    \centering
    \includegraphics[width=0.9\columnwidth]{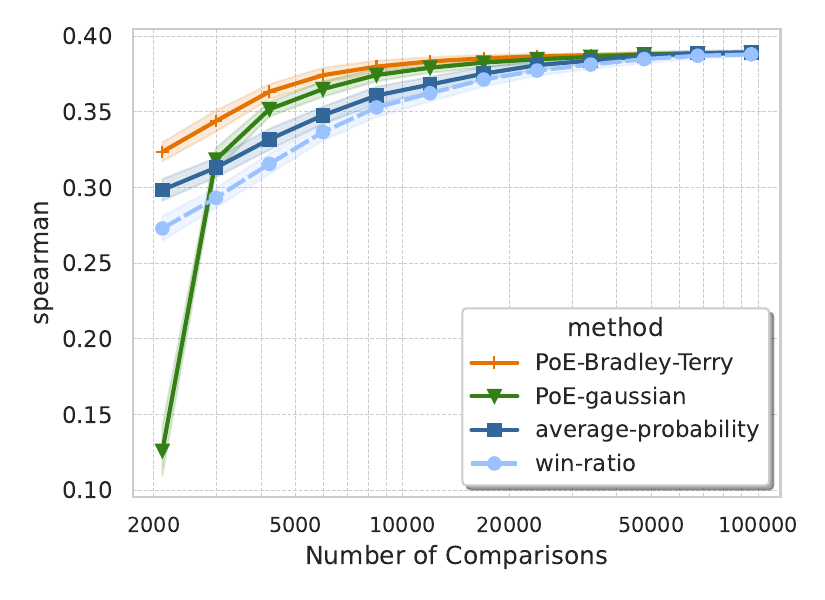}
    \vspace{-4mm}
    \caption{Mistral-7B, HANNA \texttt{COH}}
    \vspace{-4mm}
    \label{fig:efficiency_cruves_cmcqrd_hanna}
\end{figure}
\begin{figure}[h]
    \centering
    \includegraphics[width=0.9\columnwidth]{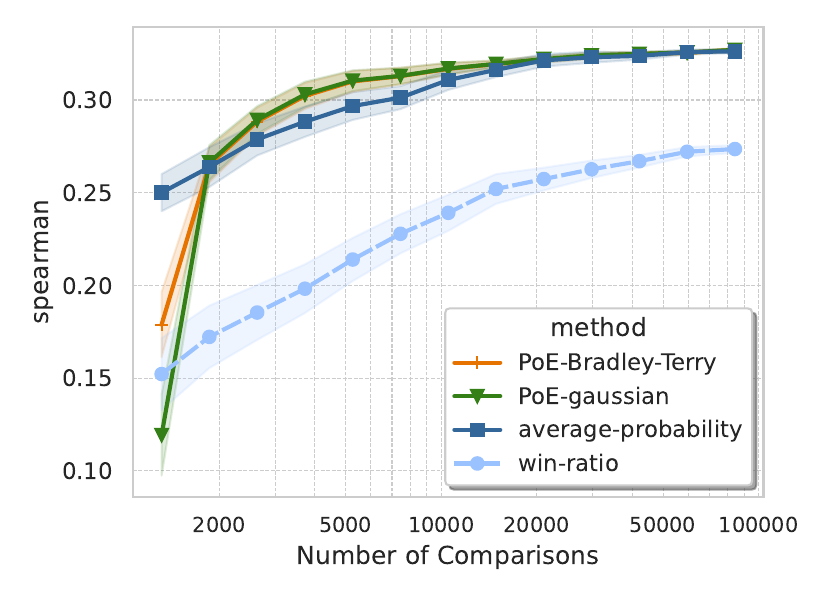}
    \vspace{-4mm}
    \caption{Llama2-13B, CMCQRD \texttt{DIF}}
    \label{fig:efficiency_cruves_cmcqrd_hanna}
    \vspace{-2mm}
\end{figure}

\begin{table*}[h]
    \centering
    \small
    \renewcommand\tabcolsep{4pt}
    \begin{tabular}{lc cc cc cc cc}
        \toprule
       &     & \multicolumn{2}{c}{CMCQRD \texttt{DIF}}  & \multicolumn{2}{c}{HANNA \texttt{COH}}  & \multicolumn{2}{c}{HANNA \texttt{CMP}} & \multicolumn{2}{c}{HANNA \texttt{SUR}} \\
        system & $K$ & avg-prob & PoE-BT & avg-prob & PoE-BT & avg-prob & PoE-BT & avg-prob & PoE-BT \\
        \cmidrule(lr){1-2} \cmidrule(lr){3-4} \cmidrule(lr){5-6} \cmidrule(lr){7-8} \cmidrule(lr){9-10}
        \multirow{4}{*}{Llama2-7B}
        & $5N$ & 31.9 & 33.4 & 39.2 & 41.3 & 45.7 & 47.9 & 32.8 & 34.1\\ 
        & $10N$ & 33.8 & 34.4 & 40.3 & 41.4 & 46.9 & 48.2 & 33.6 & 34.3\\ 
        & $20N$ & 34.8 & 35.0 & 41.1 & 41.6 & 47.6 & 48.3 & 34.1 & 34.5\\ 
        & $50N$ & 35.3 & 35.3 & 41.4 & 41.6 & 48.0 & 48.3 & 34.4 & 34.5\\ 
        \cmidrule(lr){1-2} \cmidrule(lr){3-4} \cmidrule(lr){5-6} \cmidrule(lr){7-8} \cmidrule(lr){9-10} 
        \multirow{4}{*}{Llama2-13N}
        & $5N$ & 30.0 & 31.2 & 39.9 & 41.3 & 51.7 & 54.6 & 34.6 & 36.9\\ 
        & $10N$ & 31.5 & 31.9 & 41.2 & 41.8 & 53.4 & 54.9 & 36.0 & 37.2\\ 
        & $20N$ & 32.2 & 32.3 & 41.8 & 41.9 & 54.3 & 55.1 & 36.8 & 37.5\\ 
        & $50N$ & 32.6 & 32.6 & 42.1 & 42.1 & 54.9 & 55.1 & 37.2 & 37.6\\ 
        \cmidrule(lr){1-2} \cmidrule(lr){3-4} \cmidrule(lr){5-6} \cmidrule(lr){7-8} \cmidrule(lr){9-10} 
        \multirow{4}{*}{Mistral-7B}
        & $5N$ & 38.9 & 40.7 & 36.6 & 38.3 & 47.3 & 49.9 & 24.2 & 25.5\\ 
        & $10N$ & 40.7 & 41.1 & 37.9 & 38.6 & 49.0 & 50.6 & 25.3 & 26.0\\ 
        & $20N$ & 41.1 & 41.2 & 38.7 & 38.8 & 50.1 & 50.9 & 25.9 & 26.2\\ 
        & $50N$ & 41.2 & 41.2 & 38.9 & 38.9 & 50.7 & 51.0 & 26.0 & 26.1\\ 
        \bottomrule
    \end{tabular}
    \caption{Spearman correlations for CMCQRD and HANNA for specific attributes. $K\!\in\!\{5N, 10N, 20N, 50N\}$ is the total number of symmetric comparisons made, e.g., $5N$ refers to each sample being in 5 comparisons. }
    \label{tab:poe_cmcqrd_hanna_poebt_v_avg_prob}
    \vspace{-0.3cm}
\end{table*}

The previous experiments demonstrated that the PoE framework yields significant performance boosts in efficient settings. However, for the analyzed datasets, $N$ is 16 and 6, and though PoE can reduce the number of LLM calls, it is still feasible to run all $O(N^2)$ comparisons. This section now evaluates CMCQRD and HANNA, where $N\!\!=\!\!1056$ and $N\!\!=\!\!658$ respectively. Table \ref{tab:poe_cmcqrd_hanna_poebt_v_avg_prob} presents performance when using $\alpha \cdot N$ comparisons, where it's observed that PoE-BT achieves consistently better performance than the average probability across all models and datasets. Faster convergence is observed for PoE-BT, with the average performance difference between 5 and 50 comparisons per item 0.8 SCC apart, while it is 2.5 SCC for the average probability. Note that evaluation was only conducted for Llama2 and Mistral due to FlanT5's maximum token length of 512.

Figure \ref{fig:efficiency_cruves_cmcqrd_hanna} illustrates the full efficiency curves for several models and attributes. We observe that PoE-BT typically performs best, and though PoE-g often performs similarly to PoE-BT, in very low information regions PoE-g can have poor correlations. In all cases, the PoE methods appear to mostly converge to their solution within $10 \cdot N$ comparisons, significantly fewer than $N(N\!-\!1)$.

\vspace{-1mm}
\subsection{Non-Symmetric Comparions}
\label{ssec:unbalanced}
\begin{figure}[t]
    \centering
    \includegraphics[width=0.95\columnwidth]{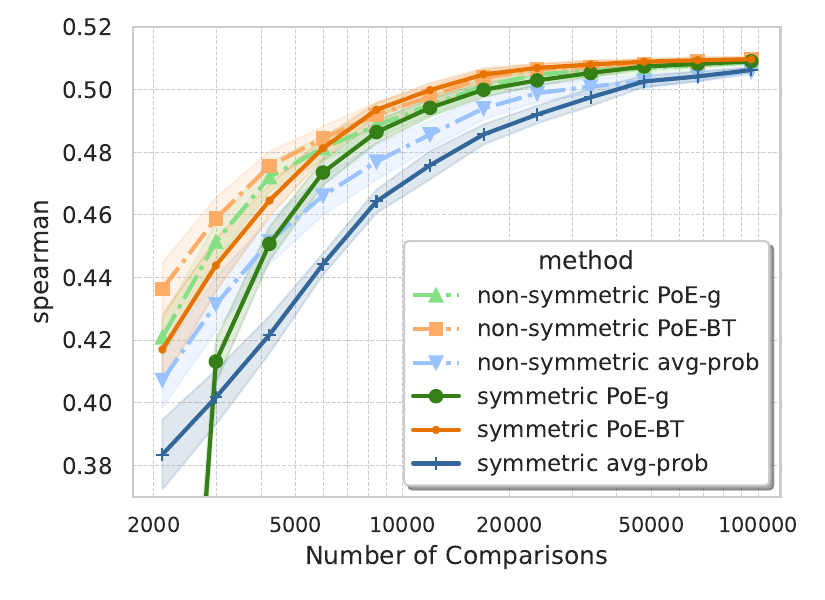}
    \captionsetup{format=plain, justification=centering}    
    \vspace{-4mm}
    \caption{Mistral-7B, HANNA \texttt{COH}, \\ symmetric vs non-symmetric}
    \label{fig:unbalanced_vs_balanced}
    \vspace{-4mm}
\end{figure}

Previously, to minimize the influence of positional bias and model inconsistency, both permutations of any comparison were evaluated. Although this reduces bias, one may gain more information by having a more diverse set of comparisons. Mistral-7B has minimal positional bias with $E[p_{ij}]\!=\!0.51$, while Llama-7B has considerable bias with $E[p_{ij}]\!=\!0.78$. To investigate whether symmetry is required, we look at performance of the non-symmetric set-up for Mistral-7B and Llama-7B (shown in Appendix Figure \ref{fig:appendix_unbalanced}). For Llama2-7B, the debiased expert yields large performance gains while for Mistral-7B, the debiasing parameter has little influence, as expected since $\gamma$ will be near 0. Note that, although Llama2-7B is more biased, it has better judgement capabilities and achieves better correlations, though the debiasing parameter is required. Figure \ref{fig:unbalanced_vs_balanced} compares non-symmetric debiased performance with symmetric performance and illustrates that the two perform similarly, albeit with slightly different characteristics. Non-symmetric often does better in the low number of comparisons region, symmetric sometimes marginally better after, and performance is similar when more comparisons are made. Results for other models and attributes are presented in Appendix \ref{ssec:app_sym_vs_nonsym}.  

\section{Conclusions}

Comparative assessment using LLMs has been shown to be effective for text assessment. This paper investigates framing the scoring process within a Product of Experts framework, where the comparison information (including model confidence) can be easily combined to determine a set of scores that effectively capture text quality. This enables comparative assessment to not suffer from slow convergence rates, as now only a subset of the possible comparisons is used to predict the scores, but maintain the performance from when using the full set of comparisons. Further, using Gaussian experts yields a closed-form solution and provides a basis for deriving a greedy-optimal set of comparisons. The paper demonstrated the effectiveness of the approach on multiple different standard NLG evaluation datasets, such as SummEval and TopicalChat, as well as for large datasets where $N\!>\!500$, which led to substantial savings in computation against standard methods. 

\section{Limitations}
\label{ssec:limitations}
The LLM comparisons can depend largely on the selected prompts used and the process used to extract probabilities. We chose simple prompts but did not investigate the impact of prompt sensitivity or how well the approach holds when weaker/stronger prompts are used. With the zero-shot nature and the consistent observed performance boosts, our method is likely to remain effective in such settings, but this was not verified. Another limitation is that when optimizing the BT experts, one can apply a soft-variant of Zermello to quickly optimise the PoE-Bradley-Taylor approach. However, when the bias term is introduced, soft-zero method cannot be applied, and optimization of the solution is significantly slower. Nonetheless, since the main computational costs are associated with LLM calls, this is not a significant drawback. Lastly, our method is effective only when soft LLM probabilities are available, and for APIs where probabilities are not available, and one can only sample binary decisions, the method is less effective.

\section{Acknowledgements}
This paper reports on research supported by Cambridge University Press \& Assessment (CUP\&A), a department of The Chancellor, Masters, and Scholars of the University of Cambridge. This research is further supported by the EPSRC (The Engineering and Physical Sciences Research Council) Doctoral Training Partnership (DTP) PhD studentship and Gates Cambridge Trust.

\section{Ethical Statement}
\label{ssec:ethical_statement}
Our paper addresses the cases of using more efficient use of LLMs when being used for NLG assessment. Although our work makes automatic assessment more practical and applicable to more settings, overly relying on automatic assessment may yield unintended consequences, especially when models have implicit biases that may discriminate against certain styles. Therefore as well as using automatic evaluation as useful metrics for text quality, it is useful to maintain human evaluation to ensure that systems do not unfairly penalize particular styles or properties which in general may be fine for the task. 

\bibliography{acl_latex}

\appendix

\section{Additional Theory for the Product of Expert Framework}
\label{sec:appendix_theory}

\subsection{Structure of $\bf{\tilde{W}}$ Matrix}
\label{sec:W_example}
The paper discussed the comparison matrix ${\bf \tilde{W}} \!\in\! R^{(K\!+\!1) \times N}$, where each row represents the particular comparison being considered. It was discussed how for the $k^{\text{th}}$ comparison between $i$ and $j$, ${\bf W}_{ki} \!=\! 1$, ${\bf W}_{kj} \!=\! -1$, and ${\bf W}_{km} \!=\! 0 \;\; \forall m \neq i, j$. Further, an extra row was prepended to ${\bf W}$ adding constraints on the first score,  forming ${\bf \tilde{W}}$ and ensuring the corresponding matrix is not defective. To illustrate the structure of ${\bf \tilde{W}}$, consider the case where one has 4 elements $x_{1:4}$ and all possible comparisons are considered, 
\begin{eqnarray}
  {\bf \tilde{W}} = \left[
    \begin{array}{c c c c}
      1 & 0 & 0 & 0  \\
      1 & -1 & 0 & 0 \\
      1 & 0 & -1 & 0 \\§
      1 & 0 & 0 & -1 \\
      0 & 1 & -1 & 0 \\
      0 & 1 & 0 & -1 \\
      0 & 0 & 1 & -1 \\ 
    \end{array}
    \right]
\end{eqnarray}

\subsection{Structure of ${\tilde{\bf W}}^\transpose {\tilde{\bf W}}$ Matrix}
In the Gaussian-Products of Experts, the variance was shown to be directly related to the matrix ${\bf \tilde{W}}^\transpose {\bf \tilde{W}}$. For the full comparison case previously considered, this would yield a matrix of the form, 
\begin{eqnarray}
    {\tilde{\bf W}}^\transpose{\tilde{\bf W}} =
  \left[
    \begin{array}{c c c c}
      4 & -1 & -1 & -1 \\
      -1 & 3 & -1 & -1 \\
      -1 & -1 & 3 & -1 \\
      -1 & -1 & -1 & 3 \\
    \end{array}
    \right]
\end{eqnarray}
Let ${\bf \tilde{A}} = {\bf \tilde{W}}^\transpose {\bf \tilde{W}}$. For any set of selected comparisons, ${\bf \tilde{A}}_{ij} = {\bf \tilde{w}}_i \cdot {\bf \tilde{w}}_j$. Therefore by taking into account the structure of $\tilde{W}$, it's easily shown that the diagonal elements represent the number of comparisons the element has been involved in, while the off-diagonal elements are -1 if the comparison is made, 
\begin{align}
    {\bf \tilde{A}}_{kk} &= \sum_i 1(x_k \in \mathcal{C}_i) \\ 
    {\bf \tilde{A}}_{ij} &= 
    \begin{cases} 
        -1 & \text{if } (x_i, x_j) \in \mathcal{C}_K, \\
        \; 0 & \text{otherwise}.
    \end{cases}
\end{align}
This means that for the full comparison matrix, irrespective of $N$, the matrix ${\tilde{\bf W}}^\transpose{\tilde{\bf W}}$ will have the form,
\begin{eqnarray*}
{\tilde{\bf W}}^\transpose{\tilde{\bf W}} =
\left[
\begin{array}{c c c c c}
   N & -1  & -1   & \ldots & -1 \\
  -1 & N\!-\!1 & -1   & \ldots & -1 \\
  -1 & -1  & N\!-\!1  & \ldots & -1\\
  \vdots & \vdots & \vdots & \ddots & \vdots\\
  -1 & -1  & -1   & \ldots & N\!-\!1 \\
\end{array}
\right]
\end{eqnarray*}

\subsection{Equivalence of Gaussian PoE Solution with Average Probability}
\label{ssec:poe-g-avg-prob-equivalence}
Given the structure of ${\tilde{\bf W}}^\transpose{\tilde{\bf W}}$, when considering the full-comparison set-up, the inverse is given by,
\begin{eqnarray*}
    &\left(\!{\tilde{\bf W}}^\transpose{\tilde{\bf W}}\!\right)^{-1} \!\!\!\!\!\!=\!\!
    \left[
    \begin{array}{c c c c c}
      1 & 1 & 1 & \ldots & 1 \\
      1 & 1\!+\!\frac{2}{N} & 1\!+\!\frac{1}{N} & \ldots & 1\!+\!\frac{1}{N} \\
      1 & 1\!+\!\frac{1}{N} & 1\!+\!\frac{2}{N}  & \ldots & 1\!+\!\frac{1}{N}\\
      \vdots & \vdots & \vdots & \ddots & \vdots\\
      1 & 1\!+\!\frac{1}{N} & 1\!+\!\frac{1}{N} & \ldots & 1\!+\!\frac{2}{N} \\
    \end{array}
    \right]
\end{eqnarray*}
\begin{eqnarray*}
     = \frac{N+1}{N}\left[
    \begin{array}{c c c c c}
      1 & 1 & 1 & \ldots & 1 \\
      1 & 1 & 1 & \ldots & 1 \\
      1 & 1 & 1  & \ldots & 1\\
      \vdots & \vdots & \vdots & \ddots & \vdots\\
      1 & 1 & 1 & \ldots & 1 \\
    \end{array}
    \right] \qquad \qquad \\
  \quad \qquad \qquad + \frac{1}{2N}
  \left[
  \begin{array}{c c c c c}
    -1 & -1 & -1 & \ldots & -1 \\
    -1 &  1 & 0 & \ldots & 0 \\
    -1 &  0 & 1  & \ldots & 0\\
    \vdots & \vdots & \vdots & \ddots & \vdots\\
    -1 & 0 & 0 & \ldots & 1 \qquad\\
  \end{array}
  \right] 
\end{eqnarray*}

For the Gaussian PoE with linear mean and constant Gaussian assumptions, the solution was shown to be of form ${\bf \hat{s}} = \alpha \cdot ({\bf \tilde{W}}^{\transpose} {\bf \tilde{W}})^{-1} {\bf \tilde{W}} \bm{\tilde{\mu}}$. By noting that $\bm{\tilde{\mu}}$ represents the LLM probabilities for each comparative decision, we observe that ${\bf \tilde{W}} \bm{\tilde{\mu}}$ simply represents the sum of probabilities for all comparisons that each element has been a part of. 
Therefore, the above equation shows that the solution will be a constant shift of the average probability for any particular sample.

\subsection{Form of the Gaussiam PoE Score Distribution}
\label{sec:rearranging_gaussian}
Given ${\tt p} ({ \bf Ws}| \mathcal{C}_{1:K}) = \mathcal{N}\left({\bf Ws}; \bm{\tilde{\mu}}, {\bf \tilde{\Sigma}} \right)$, to determine ${\tt p} ({ \bf s}| \mathcal{C}_{1:K})$ one can expand the expression and isolate all terms that have an ${\bf s}$, yielding,
\begin{align}
    &{\tt p} ({ \bf Ws}| \mathcal{C}_{1:K}) \\
    =& \mathcal{N}\!\left({\bf Ws}; \bm{\tilde{\mu}}, {\bf \tilde{\Sigma}} \right) \\
    \propto& \exp\left( \shortdash\frac{1}{2} \; ({\bf W s}-{\bm{\tilde{\mu}}})^\transpose {\bf \tilde{\Sigma}}^{-1} ({\bf W s}-{\bm{\tilde{\mu}}}) \right) \\
    \propto& \exp\left(\!\shortdash \frac{1}{2} \left( {\bf s}^\transpose {\bf W}^\transpose {\bf \tilde{\Sigma}}^{-1} {\bf W s} + 2 {\bf s}^\transpose {\bf W}^\transpose {\bf \tilde{\Sigma}}^{-1}  \tilde{\bm{\mu}} \right) \right) \notag
\end{align}
As the distribution over scores will be Gaussian, ${\tt p}({\bf s}| C_{1:K}) \sim \mathcal{N}({\bf s}; {\bm \mu^*}, {\bf \Sigma^*})$, one can equate coefficients to derive the form used in the paper,
\begin{align}
    {\bf \tilde \Sigma}^*_s &= ({\bf \tilde{W}}^\transpose {\bf \tilde \Sigma}^{-1} {\bf \tilde{W}})^{-1} \\
    {\bm{\mu}_s^*} &= ({\bf \tilde{W}}^\transpose {\bf \tilde \Sigma}^{-1} {\bf \tilde{W}})^{-1} {\bf \tilde{W}}^\transpose {\bf \tilde \Sigma}^{-1} \bm{\tilde{\mu}} 
\end{align}
Which has pdf,
\begin{equation}
    \frac{1}{(2\pi)^{N/2} |\mathbf{\tilde{\Sigma}}|^{1/2}} \exp\left(\! \shortdash \frac{1}{2} ({\bf s} \shortdash \bm{\mu}_s^*)^\transpose \mathbf{{\bf \Sigma^*}}^{-1} ({\bf s} \shortdash \bm{\mu}_s^*)\right) \notag
\end{equation}
The maximum probability scores will be at the mean, ${\bf s}=\bm{\mu}_s^*$, which has a probability of,
\begin{align}
    & \frac{1}{(2\pi)^{N/2} {\tt det}\left(({\bf \tilde{W}}^\transpose {\bf \tilde \Sigma}^{-1} {\bf \tilde{W}})^{-1}\right)^{1/2}} \\
    =& \frac{\sqrt{{\tt det} ({\bf \tilde{W}}^{\transpose} {\bf \tilde \Sigma}^{-1} {\bf \tilde{W}}) }}{ (2 \pi)^{N/2}}
\end{align}
For the linear Gaussian, where it is assumed that ${\bf \tilde \Sigma} = \sigma^2 {\bf I}$, this can be reduced to, 
\begin{align}
    {\tt p}({ \bf s} \smallequal \bm{\mu}_s^*|\mathcal{C}_{1:K}) = \frac{\sqrt{{\tt det} ({\bf \tilde{W}}^{\transpose} {\bf \tilde{W}}) }}{ (2 \pi \sigma^2)^{N/2} }
\end{align}

% \begin{align}
%     {\tt p}(\bm{\tilde{\mu}}|\mathcal{C}_{1:K}) = \frac{1}
% \end{align}

\subsection{Laplace's Approximation of Bradley-Terry}
\label{sec:laplace_approx}
As discussed in the paper, the soft Bradley-Terry model score distribution ${\tt p}(s_{1:N}|\mathcal{C}_{1:K})$ has form, 
\begin{align}
    \frac{1}{Z} \!\!\! \prod_{i,j \in \mathcal{C}_{1:K}} \!\!\! \sigma(s_i \shortdash s_j)^{p_{ij}} (1\shortdash \sigma(s_i \shortdash s_j))^{1 \shortdash p_{ij}}
\end{align}
Which can be difficult to analyze. Therefore, one can apply a Laplace approximation, which approximates the original distribution as a Gaussian, ${\tt p}({\bf s}|\mathcal{C}_{1:K}) \approx \tilde{\tt p}({\bf s}|\mathcal{C}_{1:K}) =  \mathcal{N}\!\left({\bf s}; \; \hat{{\bf s}}, \; {\bf A} \right)$,  
\begin{align*}
    \ln \tilde{\tt p}({\bf s}|\mathcal{C}_{1:K}) \!=\! \ln {\tt p}(\hat{{\bf s}}|\mathcal{C}_{1:K}) \shortdash \frac{1}{2} ({\bf s} \shortdash \hat{{\bf s}})^\transpose {\bf A}^{-1} ({\bf s} \shortdash \hat{{\bf s}})
\end{align*}
where $\hat{{\bf s}}=[\hat{s}_{1:N}]$ is the maximum probability estimate of the original distribution and ${\bf A}^{-1}$ is the inverse covariance matrix. To determine the values of ${\bf A}^{-1}$, one notes that the above equation is equivalent to a Taylor expansion of the log-distribution of the scores at $\hat{{\bf s}}$, such that 
\begin{equation}
    \left. {\bf A}^{-1}_{ij} = - \frac{\partial^2}{\partial s_i \partial s_j} \ln {\tt p}(s_{1:N}|\mathcal{C})\right|_{s_{1:N} = \hat{s}_{1:N}}
\end{equation}
By simplifying the form of score distribution using soft Bradley-Terry experts,
\begin{align}
    {\tt p}(s_{1:N}|\mathcal{C}) = \frac{1}{Z} \prod_{i,j \in \mathcal{C}} \frac{\exp(p \cdot(s_i - s_j))}{1 + \exp(s_i - s_j)}
\end{align}
and differentiating the log distribution, it can be shown that the diagonal elements, ${\bf A}^{-1}_{kk}$, and off-diagonal elements ${\bf A}^{-1}_{km}$, can be calculated as, 
% \begin{align}
%     {\bf A}^{-1}_{kk} \! &= \!\! \sum_{i, j \in \mathcal{C}} \!\! {\big [}\mathbbm{1}(i \shortequal k) \shortplus \mathbbm{1}(j \shortequal k){\big ]} \! \cdot  \! \sigma(\hat{s}_i \shortdash \hat{s}_j) \! \cdot \! \sigma(\hat{s}_j \shortdash \hat{s}_i) 
%     \label{eq:7_laplacian_bt_diag}
%     {\bf A}^{-1}_{km} &= -1 \cdot {\Big (} \! \sum_{i, j \in \mathcal{C}} \!\! {\big [} \mathbbm{1}(i \!=\! k) \mathbbm{1}(j \!=\! m) + \mathbbm{1}(j \!=\! k) \mathbbm{1}(i \!=\! m) {\big ]} \cdot \sigma(\hat{s}_i \shortdash \hat{s}_j) \cdot \sigma(\hat{s}_j \shortdash \hat{s}_i) {\Big )}
%     \label{eq:7_laplacian_bt_non_diag}
% \end{align}
\begin{eqnarray*}
    \lefteqn{{\bf A}^{-1}_{kk} \! = \!\! \sum_{i, j \in \mathcal{C}} \!\! {\big [}\mathbbm{1}(i \shortequal k) \shortplus \mathbbm{1}(j \shortequal k){\big ]} \! \cdot  \! \sigma(\hat{s}_i \shortdash \hat{s}_j) \! \cdot \! \sigma(\hat{s}_j \shortdash \hat{s}_i)} \\
    \lefteqn{{\bf A}^{-1}_{km} = -1 \cdot {\Big (} \!\! \sum_{i, j \in \mathcal{C}} {\big [} \mathbbm{1}(i \shortequal k) \mathbbm{1}(j \shortequal m)} \\
    &&
     + \mathbbm{1}(j \shortequal k) \mathbbm{1}(i \shortequal m) {\big ]} \cdot \sigma(\hat{s}_i \shortdash \hat{s}_j) \cdot \sigma(\hat{s}_j \shortdash \hat{s}_i) {\Big )}
\end{eqnarray*}

\subsection{Laplace Comparison Selection}
\label{sec:laplace_selection}

\noindent For the Laplacian approximation of the soft Bradley-Terry Experts, the inverse covariance matrix has elements as described in the above Equations. Hence, one can note that adding a further comparison (i, j) has influence on ${\bf A}$,
\begin{equation*}
    \left(\!{{\bf A}^{(k+1)}}\right)^{-1} \!\!\!\!\!=\! \left({{\bf A}^{(k)*}}\! \right)^{-1} + \sigma(\hat{s}_i \shortminus \hat{s}_j) \cdot \sigma(\hat{s}_j \shortminus \hat{s}_i) \cdot {\bf r} {\bf r}^{\transpose}
\end{equation*}
Therefore when maximizing the determinant, one can similarly demonstrate that,
\begin{align*}
    &= {\tt det}\left( \left({{\bf A}^{(k+1)}}\right)^{-1} \right) \\
    &= {\tt det}\left(\left({{\bf A}^{(k)*}}\right)^{-1} + \sigma(\hat{s}_i \shortminus \hat{s}_j) \cdot\sigma(\hat{s}_i \shortminus \hat{s}_j) \cdot {\bf r} {\bf r}^{\transpose}\right) \\
    &= 
    {\tt det}\left(\left({{\bf A}^{(k)*}}\right)^{-1}\right) \cdot {\bigg (}\\
    &\qquad \left(1 + \sigma(\hat{s}_i \shortminus \hat{s}_j) \cdot \sigma(\hat{s}_j \shortminus \hat{s}_i) \cdot {\bf r}^\transpose {{{\bf A}^{(k)*}} {\bf r}}\right) {\bigg )}
\end{align*}
Which is equivalent to maximizing, 
\begin{align*}
    \hat{i}, \hat{j} &=\argmax_{i, j} \; \sigma(\hat{s}_i \shortminus \hat{s}_j) \cdot \sigma(\hat{s}_j \shortminus \hat{s}_i) \cdot {\bf r}^\transpose {{{\bf A}^{(k)*}} {\bf r}} \\
    &= \argmax_{i, j} \; \sigma(\hat{s}_i \shortminus \hat{s}_j) \cdot \sigma(\hat{s}_j \shortminus \hat{s}_i) \cdot {\Big (} \\
    & \qquad \left({{\bf A}^{(k)*}}_{ii} + {{\bf A}^{(k)*}}_{jj} - 2 \cdot {{\bf A}^{(k)*}}_{ij} \right) {\Big )}
\end{align*}

\subsection{Efficient  Greedy Comparison Selection}
\label{sec:update_additional}
%Update of $({\bf W^\transpose W})^{-1}$ for
Assume that ${\bf \tilde{W}}^{(k)*}$ is the selected comparison matrix using $k$ comparisons. Considering an additional comparison $(i, j)$ is equivalent to adding an extra row ${\bf r} \in R^{N}$ where ${\bf r}_{i} \!=\! 1$, ${\bf r}_{j} \!=\! -1$ and ${\bf r}_{l} \!=\! 0 \;\; \forall l \neq i, j$. By noting that,
\begin{align}
    &{\tt det}\left( [{\bf \tilde{W}} ; {\bf r}]^\transpose [{\bf \tilde{W}} ; {\bf r}]\right) \\
    =& {\tt det}({\bf \tilde{W}}^{\transpose} {\bf \tilde{W}} + {\bf r} {\bf r}^\transpose ) \\
    =& {\tt det}({\bf \tilde{W}}^{\transpose} {\bf \tilde{W}}) (1 + {\bf r}^\transpose ({\bf \tilde{W}}^{\transpose} {\bf \tilde{W}})^{-1}{\bf r})
\end{align}
the next optimal comparison $(\hat{i}, \hat{j})$ is calculated as,  
\begin{equation}
    \hat{i}, \hat{j} = \argmax_{i, j} {\bf A}^{(k)*}_{ii} + {\bf A}^{(k)*}_{jj} - 2 \cdot {\bf A}^{(k)*}_{ij}  
\end{equation}
Updating ${\bf \tilde{W}}^{(k)*}$ is trivial, since considering an additional comparison $(i, j)$ is equivalent to adding an extra row ${\bf r} \in R^{N}$ to ${\bf \tilde{W}}^{(k)*}$, where ${\bf r}_{i} \!=\! 1$, ${\bf r}_{j} \!=\! -1$ and ${\bf r}_{l} \!=\! 0 \;\; \forall l \neq i, j$. Therefore 
\begin{equation}
    {\bf \tilde{W}}^{(k+1)*} = [{\bf \tilde{W}}^{(k)*}; {\bf r}]
\end{equation}
However one can also efficiently update the inverse using the Sherman-Morrison inversion lemma,  
\begin{align}
    {\bf A}^{(k+1)*} &= \left( [{\bf \tilde{W}}^{(k)*} ; {\bf r}]^\transpose [{\bf \tilde{W}}^{(k)*} ; {\bf r}] \right)^{-1} \\
    &= \left( {\bf \tilde{W}}^{(k)*^\transpose} {\bf \tilde{W}}^{(k)*} + {\bf r} {\bf r}^\transpose \right)^{-1} \\
    % &= \left( ({\bf A}^{(k)*})^{-1} + {\bf r}^\transpose {\bf r} \right)^{-1} \\
    &= {\bf A}^{(k)*} - \frac{{\bf A}^{(k)*} {\bf r} {\bf r}^\transpose {\bf A}^{(k)*}}{1 + {\bf r}^\transpose {\bf A}^{(k)*} {\bf r}}
    % &= {\bf A}^{(k)} + \frac{\left({\bf A}^{(k)*}\right)^{-1} {\bf r} {\bf r}^\transpose \left({\bf A}^{(k)*}\right)^{-1}}{1 + {\bf r}^\transpose \left({\bf A}^{(k)*}\right)^{-1} {\bf r}}
\end{align}
Note that to initialize ${\bf \tilde{W}}$, the simplest option would be to use $N-1$ comparisons and follow a stripped diagonal matrix, e.g.
\begin{eqnarray}
  {\bf \tilde{W}} = \left[
    \begin{array}{c c c c}
      1 & 0 & 0 & 0 \\
      1 & -1 & 0 & 0 \\
      0 & 1 & -1 & 0 \\
      0 & 0 & 1 & -1 \\
    \end{array}
    \right]
\end{eqnarray}

% \subsection{Gaussian Assumptions under a Biased System}

% In Section \ref{ssec:gaussian_assumptions}, it was discussed how one could set the mean of the Gaussian Product-of-Expert model to $f_{\mu}(p) = \alpha \! \cdot \! (p - \beta)$, where $\beta = 0.50$ for an unbiased comparative assessment system. However, in a single comparative assessment call where we task the underlying LLM with predicting which candidate is better, the predicted probabilities suffer from positional bias. To accommodate this, one can either permute the options and call the LLM again to eliminate positional bias, or change the modelling structure. The latter can be done by shifting the bias term $\beta$ according to:
% \begin{align}
%     \beta = \frac{1}{K} \sum_{i, j \in C_{1:K}} p_{ij}
% \end{align}
% Using this $\beta$ will allow the Gaussian expert to accommodate for the bias that exists in the outputs of the underlying LLM, without resorting to calling the LLM for both permutations.

\subsection{Detailed Derivation of $\beta$ for the Debiased PoE-Gaussian Expert}
\label{ssec:accounting_for_bias_g}
For a given expert, ${\tt p}(s_i - s_j|p_{ij})$, and an underlying LLM which generates comparative decisions, ${\tt p}_{\text{LM}}(p_{ij})$ (assuming the underlying texts $x_i$ and $x_j$ are randomly drawn), there is an associated marginalised distribution of score differences, ${\tt p}(s_i - s_j)$. Note that as the texts are randomly drawn, they are equally likely to be drawn in either position and therefore, $\mathbb{E}[s_i - s_j]=0$. For a debiased expert ${\tt p}_{\gamma}(s_i - s_j|p_ij)$, the objective is to find the parameter $\gamma$ for the LLM that ensures that $\mathbb{E}[s_i - s_j]=0$,
\begin{align}
    & \mathbb{E}[s_i - s_j] \\
    &= \!\! \int^{\infty}_{-\infty} (s_i \vshortdash s_j) {\tt p}(s_i \vshortdash s_j) d(s_i \vshortdash s_j) \\
    &= \!\! \int^1_0 \!\!\! \int^{\infty}_{-\infty}  \!\!\!\!\! (s_i \vshortdash s_j) {\tt p}_{\gamma}(s_i \vshortdash s_j | p_{ij}) {\tt p}_{\text{LM}}(p_{ij})  d(s_i \vshortdash s_j) dp_{ij} \notag \\ 
    &= \int^1_0 \!\!{\tt p}_{\text{LM}}(p_{ij}) \!\! \int^{\infty}_{-\infty} \!\!\!\!\! (s_i \vshortdash s_j) {\tt p}_{\gamma}(s_i \vshortdash s_j | p_{ij}) d(s_i \vshortdash s_j) dp_{ij} \notag \\ 
    &= \int^1_0 \!\! {\tt p}_{\text{LM}}(p_{ij}) \cdot \mathbb{E}[s_i - s_j | p_{ij}, {\gamma}] \; dp_{ij}
\end{align}
The parameter $\gamma$ was proposed to be a simple linear shift of the score differences, such that ${\tt p}_{\gamma}(s_i \shortdash s_j | p_{ij}) = {\tt p}(s_i \shortdash s_j \! \shortdash \gamma | p_{ij})$. For the linear Gaussian, $\mathcal{N}{\big (}s_i \shortdash s_j; \alpha \! \cdot \! (p_{ij} \shortdash \beta), \sigma^2{\big )}$ this is equivalent to setting the $\beta$ parameter. The mean of the expert is $\alpha \! \cdot \! (p_{ij} \shortdash \beta)$, and therefore, 
\begin{align}
    \mathbb{E}[s_i \shortdash s_j] &= \!\! \int^1_0 {\tt p}_{\text{LM}}(p_{ij}) \cdot \mathbb{E}[s_i \shortdash s_j | p_{ij}] \; dp_{ij} \\
    &= \!\! \int^1_0 {\tt p}_{\text{LM}}(p_{ij}) \cdot \alpha \! \cdot \! (p_{ij} \shortdash \beta) \; dp_{ij} \\
    &= \alpha \left( \!\! \int^1_0 p_{ij} \; {\tt p}_{\text{LM}}(p_{ij}) \; dp_{ij} - \beta \right)
\end{align}
Which setting to zero yields $\beta = \mathbb{E}[p_{ij}] \approx \frac{1}{K}\sum_{k=1}^Kp_{ij}^{(k)}$, i.e. $\beta$ should be set to the average LLM probability.

% \subsection{Deriving $\gamma$ for the Debiased PoE-BT Expert}
% \label{ssec:accounting_for_bias_bt}
% For experts which are unstable, or for which the expectation is analytically intractable, one can instead ensure the mode of the skill difference likelihood is set to when the skill difference to 0:
% \begin{align}
%     & \frac{\partial}{\partial \gamma} \mathbb{E}_{i \neq j}\left[\log {\tt p}_{\gamma}(s_i - s_j)\right]\Big\vert_{s_i-s_j=0} \\
%     = & \frac{\partial}{\partial \gamma} \mathbb{E}_{i \neq j} \left[p_{ij} \sigma(s_i - s_j - \gamma) + (1 - p_{ij}) \left(1 - \sigma(s_i - s_j - \gamma)\right)\right] \Big\vert_{s_i-s_j=0} \\
%     = & - \mathbb{E}_{i \neq j} \left[ p_{ij} - \sigma(s_i - s_j - \gamma) \right] \Big\vert_{s_i-s_j=0} \\
%     = & - \mathbb{E}_{i \neq j} \left[ p_{ij} \right] + \sigma(-\gamma) = 0
% \end{align}
% This finally results in the following: $\gamma = -\log \left( \frac{\mathbb{E}_{i \neq j}[p_{ij}]}{1+\mathbb{E}_{i \neq j}[p_{ij}]} \right) = -\text{logit}(\mathbb{E}_{i \neq j}[p_{ij}])$.

\subsection{Deriving $\gamma$ for the Debiased PoE-BT Expert}
\label{ssec:accounting_for_bias_bt}
For experts that are unstable or for which the expectation is analytically intractable, one can instead ensure the mode of the skill difference likelihood is set to 0 when the skill difference is 0. Differentiating the expected score difference yields,
\begin{align}
       &\frac{\partial}{\partial \gamma} \mathbb{E}[\log {\tt p}_{\gamma}(s_i - s_j)] \\
       =& \frac{\partial}{\partial \gamma} \int^1_0 \log {\tt p}_{\gamma}(s_i \shortdash s_j | p_{ij}) {\tt p}_{\text{LM}}(p_{ij}) dp_{ij} \\
       =& \!\! \int^1_0 \!\! {\tt p}_{\text{LM}}(p_{ij}) \frac{\partial}{\partial \gamma} {\Big (} \log {\tt p}_{\gamma}(s_i \shortdash s_j | p_{ij}) {\Big )} dp_{ij}
\end{align}
The probabilistic Bradley-Terry accounting for bias has form, 
\begin{align}
   {\tt p}_{\gamma}(s_i - s_j | p_{ij}) = \frac{1}{Z_{ij}} \cdot \frac{e^{p_{ij} \cdot (s_i - s_j - \gamma)}}{1 + e^{(s_i - s_j - \gamma)}} 
\end{align}
which when differentiated yields, 
\begin{align}
& \frac{\partial}{\partial \gamma} \log {\tt p}(s_i - s_j| p) \\
=& \frac{\partial}{\partial \gamma} {\big (} p_{ij}\cdot(s_i \shortdash s_j \shortdash \gamma) \shortdash \log(1 + e^{s_i \shortdash s_j \shortdash \gamma}) {\big )} \\
=& - p_{ij} + \frac{e^{s_i \shortdash s_j \shortdash \gamma}}{1+ e^{s_i \shortdash s_j \shortdash \gamma}}
\end{align}

Evaluating the integral at $s_i-s_j=0$, 
\begin{align}
    &\left.\frac{\partial}{\partial \gamma} \mathbb{E}[\log {\tt p}_{\gamma}(s_i - s_j)] \right|_{s_i \! - \! s_j \! = \! 0} \\
    =& \int^1_0 {\tt p}_{\text{LM}}(p_{ij}) \left(p_{ij} + \frac{e^{-\gamma}}{1+e^{-\gamma}}\right) dp_{ij}
\end{align}
setting to zero yields, $\gamma \!=\! -1 \!\cdot\! \log \left( \frac{\mathbb{E}[p_{ij}]}{1+\mathbb{E}[p_{ij}]} \right) = -\text{logit}(\mathbb{E}[p_{ij}]) \approx \text{logit}\left(\frac{1}{K}\sum_{k=1}^Kp_{ij}^{(k)}\right)$

\section{Experimental Details}
\subsection{Prompts}
\label{ssec:prompts}
Table \ref{tab:prompts} shows examples of the prompts used for generating comparative decisions (other prompts for other attributes were of similar style). For a particular dataset and attribute, all models are provided with the same simple prompts, which were the only prompts used for experiments. No prompt engineering was done, matching situations where one doesn't have access to labels to evaluate systems. 

\subsection{Computation Resources}
\label{ssec:compute}
All experiments were run on L40 machines, where evaluation was parallelised over 4 machines. Each SummEval attribute took a 1 L40 GPU hours for Llama2-7b, Mistral-7B, and FlanT5-3B (despite being smaller, FlanT5 is float32 and hence not faster) while Llama2-13B took 2 hours and FlanT5-11B took 2.5 hours. For each attribute of HANNA, performing 200,000 comparisons required 8/8/9/15/21 GPU hours for Llama2-7B/Mistral-7B/FlanT5-3B/Llama2-13B/FlanT5-11B.  For CMCQRD performing 200,000 comparisons required 8/8/9/15/21 GPU hours for Llama2-7B/Mistral-7B/FlanT5-3B/Llama2-13B/FlanT5-11B. All TopicalChat experiments could be run in under 30 minutes.

\subsection{Model and Dataset Licences}
\label{sec:license}

\textbf{Model Licenses}: LLaMA-2-7B-chat and LLaMA-2-13B-chat \cite{touvron2023llama} use a LLaMA-2 license. Mistral-7B-Instruct-v0.2 uses an Apache-2.0 license. Similarly, FlanT5-3B and FlanT5-11B use an Apache-2.0 license. \vspace{2mm}

\noindent \textbf{Dataset Licenses}: SummEval \cite{fabbri2021summeval} uses an MIT License. TopicalChat \cite{mehri2020usr} uses the MIT License. Hanna \cite{chhun2022human} uses an MIT License. CMCQRD \cite{mullooly2023cambridge} uses its own license.

% Links to the following licenses that apply to the models are provided below:
% \begin{itemize}
%     \item Llama2: \url{https://huggingface.co/meta-llama/Llama-2-7b-chat-hf/blob/main/LICENSE.txt}
%     \item Apache-2.0: \url{https://www.apache.org/licenses/LICENSE-2.0}
%     \item MIT License: \url{https://choosealicense.com/licenses/mit/}
%     \item CMCQRD: \url{https://englishlanguageitutoring.com/datasets/cambridge-multiple-choice-questions-reading-dataset}
% \end{itemize}

\onecolumn
\begin{table}[t]
    \centering
    \small
    \begin{tabular}{ccp{9.5cm}}
        \toprule
        dataset & score & \multicolumn{1}{c}{prompt} \\
        %\cmidrule(lr){1-2} \cmidrule(lr){3-3} 
        \midrule
         \multirow{2}{*}{SummEval} & \multirow{2}{*}{\texttt{COH}} & Article: <context>\n \n Summary A: <A> \n \n Summary B: <B> \n \n Which Summary is more coherent, Summary A or Summary B? \\
        \cmidrule(lr){1-2} \cmidrule(lr){3-3} 
         \multirow{2}{*}{SummEval} & \multirow{2}{*}{\texttt{CON}} & Article: <context> \n \n Summary A: <A> \n \n Summary B: <B> \n \n Which Summary is more consistent to the article, Summary A or Summary B? \\
        \midrule
         \multirow{2}{*}{TopicalChat} & \multirow{2}{*}{\texttt{CNT}} & Dialogue: <context> \n \n Response A: <A> \n \n Response B: <B> \n \n Which Response continues the dialogue better, Response A or Response B?\\
        \cmidrule(lr){1-2} \cmidrule(lr){3-3} 
         \multirow{2}{*}{TopicalChat} & \multirow{2}{*}{\texttt{NAT}} & Dialogue: <context> \n \n Response A: <A> \n \n Response B: <B> \n \n Which Response appears more natural, Response A or Response B?\\
        \midrule
        \multirow{2}{*}{HANNA} & \multirow{2}{*}{\texttt{SUR}} & Story A: \n <A> \n \n Story B: \n <B> \n \n Which story is more surprising, Story A or Story B? \\
        \cmidrule(lr){1-2} \cmidrule(lr){3-3} 
        \multirow{2}{*}{HANNA}  & \multirow{2}{*}{\texttt{CMP}} & Story A: \n <A> \n \n Story B: \n <B> \n \n Which story is more complex, Story A or Story B? \\
        \midrule
        \multirow{2}{*}{CMCQRD}  & \multirow{2}{*}{\texttt{DIF}} & Question A: \n <A> \n \n Question B: \n <B> \n \n Which reading comprehension question is more difficult to answer, Question A or Question B? \\
         \bottomrule
    \end{tabular}
    \vspace{1mm}
    \caption{Prompts used for prompting the LLM to make pairwise decisions between two candidate texts.}
    \label{tab:prompts}
\end{table}

\section{Additional Results}

\label{sec:extra_results}
\subsection{SummEval Pearson Performance Tables}
The main paper illustrated the context-level Spearman correlations for SummEval, which Table \ref{tab:summeval_pearson_table_app} also shows the standard deviations of. For certain applications, one may not only care about the rank ordering of the points but also the relative spacing between them, as this provides information on the predicted quality difference between any two texts. Table \ref{tab:summeval_pearson_table} therefore presents the Pearson correlations for SummEval, where similar trends to the  Spearman table are observed. 

\begin{table}[h]
    \centering
    \small
    \renewcommand\tabcolsep{4pt}
    \begin{tabular}{lrcccccc}
        \toprule
               &     & \multicolumn{3}{c}{decisions only}  & \multicolumn{3}{c}{ probabilities} \\
        system & $K$ & win-ratio & BT & PoE-g-hard & avg-prob & PoE-BT & PoE-g \\
        \cmidrule(lr){1-2} \cmidrule(lr){3-5} \cmidrule(lr){6-8}
        \multirow{2}{*}{Llama2-7B} 
        & 48  & 21.6\std{0.8} & 23.4\std{0.7} & 22.5\std{0.7} & 24.0\std{0.7} & 26.8\std{0.5} & 26.6\std{0.5} \\
        & 240 & 27.8\std{0.0} & 27.9\std{0.0} & 27.6\std{0.0} & 28.4\std{0.0} & 28.4\std{0.0} & 28.4\std{0.0} \\
        \cmidrule(lr){1-2} \cmidrule(lr){3-5} \cmidrule(lr){6-8}
        \multirow{2}{*}{Llama2-13B} 
        & 48  & 30.8\std{0.7} & 33.1\std{0.7} & 31.6\std{0.7} & 33.7\std{0.6} & 37.7\std{0.4} & 37.3\std{0.4} \\
        & 240 & 39.3\std{0.0} & 39.3\std{0.0} & 39.2\std{0.0} & 39.3\std{0.0} & 39.3\std{0.0} & 39.3\std{0.0} \\
        \cmidrule(lr){1-2} \cmidrule(lr){3-5} \cmidrule(lr){6-8}
        \multirow{2}{*}{Mistral-7B} 
        & 48  & 29.7\std{0.8} & 31.9\std{0.7} & 30.5\std{0.6} & 31.1\std{0.7} & 33.2\std{0.6} & 32.8\std{0.6} \\
        & 240 & 38.1\std{0.0} & 38.1\std{0.0} & 38.0\std{0.0} & 37.7\std{0.0} & 37.7\std{0.0} & 37.7\std{0.0} \\
        \cmidrule(lr){1-2} \cmidrule(lr){3-5} \cmidrule(lr){6-8}
        \multirow{2}{*}{FlanT5-3B} 
        & 48  & 34.1\std{0.8} & 36.6\std{0.6} & 34.9\std{0.7} & 38.4\std{0.6} & 42.6\std{0.4} & 42.4\std{0.4} \\
        & 240 & 43.6\std{0.0} & 43.6\std{0.0} & 43.4\std{0.0} & 44.3\std{0.0} & 44.3\std{0.0} & 44.3\std{0.0} \\
        \cmidrule(lr){1-2} \cmidrule(lr){3-5} \cmidrule(lr){6-8}
        \multirow{2}{*}{FlanT5-11B} 
        & 48  & 31.2\std{0.8} & 33.4\std{0.7} & 32.0\std{0.7} & 34.7\std{0.7} & 38.5\std{0.4} & 38.4\std{0.4} \\
        & 240 & 40.0\std{0.0} & 40.0\std{0.0} & 39.7\std{0.0} & 40.5\std{0.0} & 40.5\std{0.0} & 40.5\std{0.0} \\
        \bottomrule
    \end{tabular}
    \caption{Spearman Correlations for SummEval, averaged over all attributes ({\tt COH}, {\tt CON}, {\tt FLU}, {\tt REL}). $K$ is the number of comparisons made, where $K\!=\!240$ is the full set of comparisons.}
    \label{tab:summeval_pearson_table_app}
\end{table}

\begin{table}[h!]
    \centering
    \small
    \renewcommand\tabcolsep{4pt}
    \begin{tabular}{lr ccc ccc}
        \toprule
        system & $R$ & win-ratio & BT & PoE-g-hard & avg-prob & PoE-BT & PoE-g \\
        \cmidrule(lr){1-2} \cmidrule(lr){3-5} \cmidrule(lr){6-8}
        \multirow{2}{*}{Llama2-7B}  
         & 48  & 21.7\std{0.7} & 23.5\std{0.6} & 22.3\std{0.7} & 24.3\std{0.6} & 26.9\std{0.5} & 26.8\std{0.4} \\
         & 240 & 27.8\std{0.0} & 27.8\std{0.0} & 27.8\std{0.0} & 28.4\std{0.0} & 28.4\std{0.0} & 28.4\std{0.0} \\
        \cmidrule(lr){1-2} \cmidrule(lr){3-5} \cmidrule(lr){6-8}
        \multirow{2}{*}{Llama2-13B}  
         & 48  & 31.3\std{0.7} & 33.8\std{0.6} & 32.0\std{0.7} & 36.0\std{0.5} & 40.6\std{0.3} & 39.9\std{0.4} \\
         & 240 & 39.8\std{0.0} & 40.4\std{0.0} & 39.9\std{0.0} & 42.1\std{0.0} & 42.5\std{0.0} & 42.1\std{0.0} \\
        \cmidrule(lr){1-2} \cmidrule(lr){3-5} \cmidrule(lr){6-8}
        \multirow{2}{*}{Mistral-7B}  
         & 48  & 30.8\std{0.7} & 33.3\std{0.7} & 31.6\std{0.6} & 32.5\std{0.6} & 35.5\std{0.7} & 34.7\std{0.7} \\
         & 240 & 39.7\std{0.0} & 40.5\std{0.0} & 39.7\std{0.0} & 39.9\std{0.0} & 41.3\std{0.0} & 39.9\std{0.0} \\
        \cmidrule(lr){1-2} \cmidrule(lr){3-5} \cmidrule(lr){6-8}
        \multirow{2}{*}{FlanT5-3B}  
         & 48  & 34.3\std{0.8} & 37.2\std{0.7} & 35.0\std{0.7} & 42.3\std{0.5} & 48.3\std{0.3} & 47.1\std{0.3} \\
         & 240 & 44.1\std{0.0} & 45.0\std{0.0} & 44.1\std{0.0} & 49.4\std{0.0} & 50.0\std{0.0} & 49.4\std{0.0} \\
        \cmidrule(lr){1-2} \cmidrule(lr){3-5} \cmidrule(lr){6-8}
        \multirow{2}{*}{FlanT5-11B}  
         & 48  & 31.7\std{0.7} & 34.2\std{0.7} & 32.3\std{0.7} & 37.3\std{0.6} & 41.8\std{0.5} & 41.4\std{0.5} \\
         & 240 & 40.8\std{0.0} & 41.4\std{0.0} & 40.8\std{0.0} & 43.7\std{0.0} & 44.0\std{0.0} & 43.7\std{0.0} \\
        \bottomrule
    \end{tabular}
    \caption{Pearson correlations for SummEval, averaged over all attributes ({\tt COH}, {\tt CON}, {\tt FLU}, {\tt REL}). $K$ is the number of balanced comparisons made, where $K\!=\!120$ is the full set of comparisons.}
    \vspace{-3mm}
    \label{tab:summeval_pearson_table}
\end{table}

\newpage
\subsection{TopicalChat Performance Tables}
Table \ref{tab:topical_chat_spearman} and \ref{tab:topical_chat_pearson} demonstrate performance for comparative assessment when applied to dialogue evaluation. The PoE approaches continue to provide considerable performance improvements at the operating point $K\!=\!18$, albeit since $N$ is not very large ($N\!=\!6$), the full set of comparisons is only 30 comparisons and fairly feasible to compute, and so for these experiments the computational savings are less significant. 

\label{ssec:extra_results_table}
\begin{table}[h]
    \centering
    \small
    \renewcommand\tabcolsep{4pt}
    \begin{tabular}{lr ccc ccc}
        \toprule
        system & $R$ & win-ratio & BT & PoE-g-hard & avg-prob & PoE-BT & PoE-g \\
        \cmidrule(lr){1-2} \cmidrule(lr){3-5} \cmidrule(lr){6-8}
        \multirow{2}{*}{Llama2-7B}  
        & 18  & 28.4\std{1.2} & 28.9\std{1.0} & 28.7\std{1.1} & 27.7\std{1.4} & 29.7\std{0.9} & 29.5\std{1.0} \\
        & 30  & 31.5\std{0.0} & 31.6\std{0.0} & 31.6\std{0.0} & 31.5\std{0.0} & 31.5\std{0.0} & 31.5\std{0.0} \\
        \cmidrule(lr){1-2} \cmidrule(lr){3-5} \cmidrule(lr){6-8}
        \multirow{2}{*}{Llama2-13B}  
        & 18  & 37.4\std{1.1} & 38.1\std{1.1} & 37.9\std{1.0} & 38.4\std{1.2} & 40.5\std{0.8} & 40.5\std{0.9} \\
        & 30  & 41.6\std{0.0} & 41.7\std{0.0} & 41.8\std{0.0} & 41.6\std{0.0} & 41.6\std{0.0} & 41.6\std{0.0} \\
        \cmidrule(lr){1-2} \cmidrule(lr){3-5} \cmidrule(lr){6-8}
        \multirow{2}{*}{Mistral-7B}  
        & 18  & 42.8\std{1.1} & 43.3\std{0.9} & 43.2\std{1.3} & 42.8\std{1.2} & 45.3\std{1.1} & 44.8\std{1.0} \\
        & 30  & 47.4\std{0.0} & 47.2\std{0.0} & 47.7\std{0.0} & 46.9\std{0.0} & 46.9\std{0.0} & 46.9\std{0.0} \\
        \cmidrule(lr){1-2} \cmidrule(lr){3-5} \cmidrule(lr){6-8}
        \multirow{2}{*}{FlanT5-3B}  
        & 18  & 41.3\std{1.3} & 41.8\std{1.2} & 41.6\std{1.3} & 43.4\std{1.2} & 45.4\std{0.8} & 45.2\std{0.8} \\
        & 30  & 45.3\std{0.0} & 44.8\std{0.0} & 45.3\std{0.0} & 44.7\std{0.0} & 44.7\std{0.0} & 44.7\std{0.0} \\
        \cmidrule(lr){1-2} \cmidrule(lr){3-5} \cmidrule(lr){6-8}
        \multirow{2}{*}{FlanT5-11B}  
        & 18  & 51.2\std{1.2} & 52.4\std{1.1} & 51.9\std{1.1} & 53.8\std{1.1} & 56.2\std{0.8} & 56.1\std{0.8} \\
        & 30  & 57.0\std{0.0} & 56.6\std{0.0} & 56.0\std{0.0} & 58.1\std{0.0} & 58.1\std{0.0} & 58.1\std{0.0} \\
        \bottomrule
    \end{tabular}
    \caption{Spearman correlations for TopicalChat, averaged over all attributes ({\tt COH}, {\tt CNT}, {ENG}, {\tt NAT}). $K$ is the number of comparisons made, where $K\!=\!30$ is the full set of comparisons.}
    \label{tab:topical_chat_spearman}
\end{table}

\begin{table}[h!]
    \centering
    \small
    \renewcommand\tabcolsep{4pt}
    \begin{tabular}{lr ccc ccc}
        \toprule
        system & $R$ & win-ratio & BT & PoE-g-hard & avg-prob & PoE-BT & PoE-g \\
        \cmidrule(lr){1-2} \cmidrule(lr){3-5} \cmidrule(lr){6-8}
        \multirow{2}{*}{Llama2-7B}  
         & 18  & 28.5\std{1.1} & 29.4\std{0.8} & 29.1\std{1.0} & 29.1\std{1.1} & 29.4\std{0.8} & 30.2\std{0.7} \\
         & 30  & 31.6\std{0.0} & 31.6\std{0.0} & 31.6\std{0.0} & 31.5\std{0.0} & 30.7\std{0.0} & 31.5\std{0.0} \\
        \cmidrule(lr){1-2} \cmidrule(lr){3-5} \cmidrule(lr){6-8}
        \multirow{2}{*}{Llama2-13B}  
         & 18  & 37.5\std{1.1} & 38.7\std{1.0} & 38.4\std{1.0} & 40.2\std{1.0} & 41.8\std{0.5} & 41.8\std{0.6} \\
         & 30  & 41.4\std{0.0} & 41.5\std{0.0} & 41.4\std{0.0} & 42.5\std{0.0} & 42.6\std{0.0} & 42.5\std{0.0} \\
        \cmidrule(lr){1-2} \cmidrule(lr){3-5} \cmidrule(lr){6-8}
        \multirow{2}{*}{Mistral-7B}  
         & 18  & 42.0\std{1.1} & 43.2\std{0.9} & 43.0\std{1.2} & 44.4\std{1.0} & 46.1\std{0.9} & 46.1\std{0.7} \\
         & 30  & 46.4\std{0.0} & 46.3\std{0.0} & 46.4\std{0.0} & 48.1\std{0.0} & 48.4\std{0.0} & 48.1\std{0.0} \\
        \cmidrule(lr){1-2} \cmidrule(lr){3-5} \cmidrule(lr){6-8}
        \multirow{2}{*}{FlanT5-3B}  
         & 18  & 42.1\std{1.2} & 43.1\std{1.1} & 42.8\std{1.1} & 45.7\std{1.0} & 48.0\std{0.7} & 47.9\std{0.7} \\
         & 30  & 46.5\std{0.0} & 46.5\std{0.0} & 46.5\std{0.0} & 48.7\std{0.0} & 48.6\std{0.0} & 48.7\std{0.0} \\
        \cmidrule(lr){1-2} \cmidrule(lr){3-5} \cmidrule(lr){6-8}
        \multirow{2}{*}{FlanT5-11B}  
         & 18  & 51.5\std{1.2} & 53.3\std{1.0} & 52.9\std{1.0} & 56.3\std{0.9} & 58.1\std{0.6} & 58.3\std{0.6} \\
         & 30  & 57.5\std{0.0} & 57.4\std{0.0} & 57.4\std{0.0} & 59.8\std{0.0} & 59.7\std{0.0} & 59.8\std{0.0} \\
        \bottomrule
    \end{tabular}
    \caption{Pearson correlations for TopicalChat averaged over all attributes ({\tt COH}, {\tt CNT}, {ENG}, {\tt NAT}). $K$ is the number of comparisons made, where $K\!=\!30$ is the full set of comparisons.}
    \vspace{-3mm}
    \label{tab:topical_chat_pearson}
\end{table}

\newpage
\subsection{SummEval and Topical Chat Efficiency Plots}
\label{ssec:extra_efficient_plots_summeval}

\begin{figure}[h]
    \centering
    \begin{subfigure}[t]{0.33\textwidth}
         \centering
         \includegraphics[width=\textwidth]{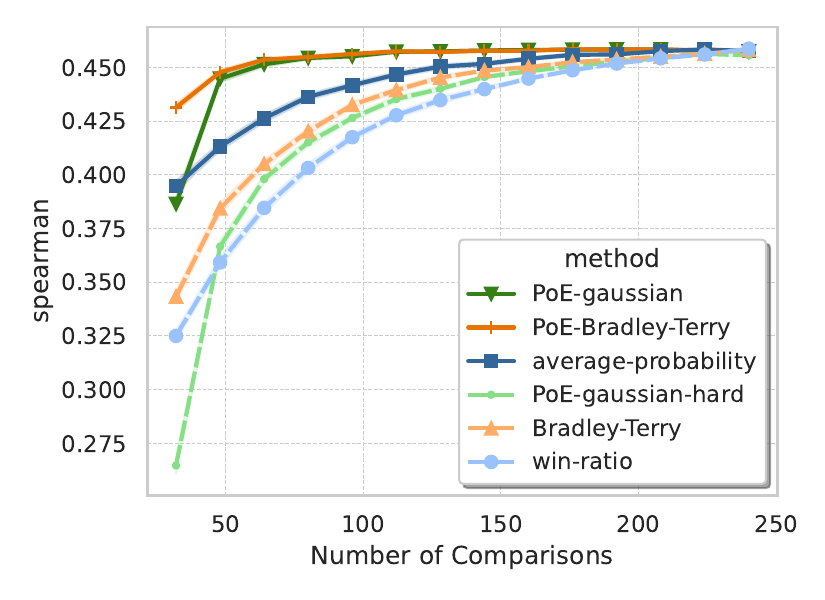}
        \caption{\smallcaption FlanT5-3B, SummEval \texttt{CON}}
    \end{subfigure}%
    ~ 
    \begin{subfigure}[t]{0.33\textwidth}
        \centering
         \includegraphics[width=\textwidth]{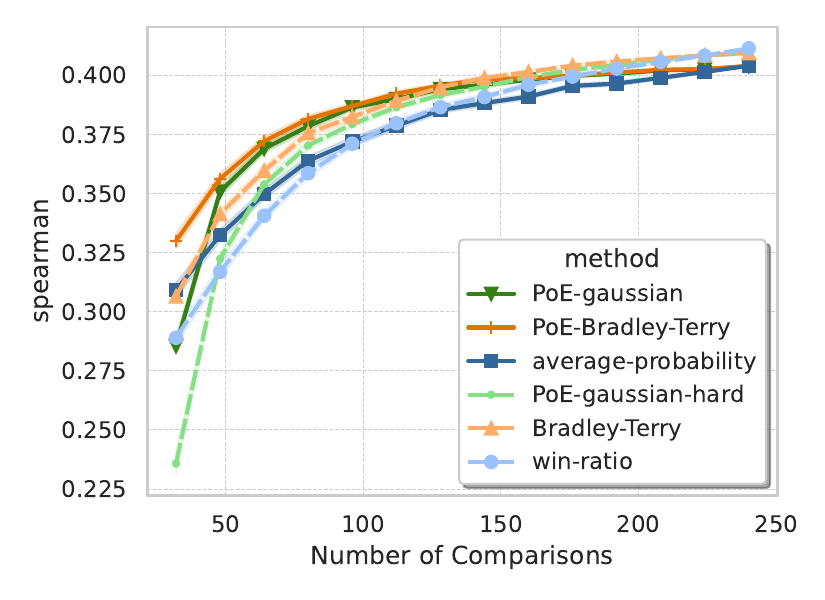}
        \caption{\smallcaption Mistral-7B, SummEval \texttt{COH}}
    \end{subfigure}%
    ~ 
    \begin{subfigure}[t]{0.33\textwidth}
        \centering
         \includegraphics[width=\textwidth]{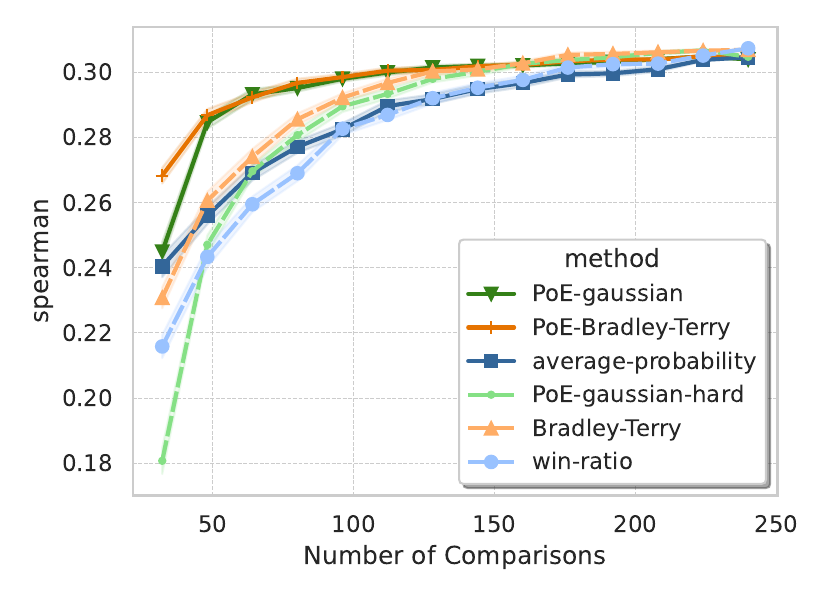}
        \caption{\smallcaption Llama-13B, SummEval \texttt{FLU}}
    \end{subfigure} \\
    \begin{subfigure}[t]{0.33\textwidth}
         \centering
         \includegraphics[width=\textwidth]{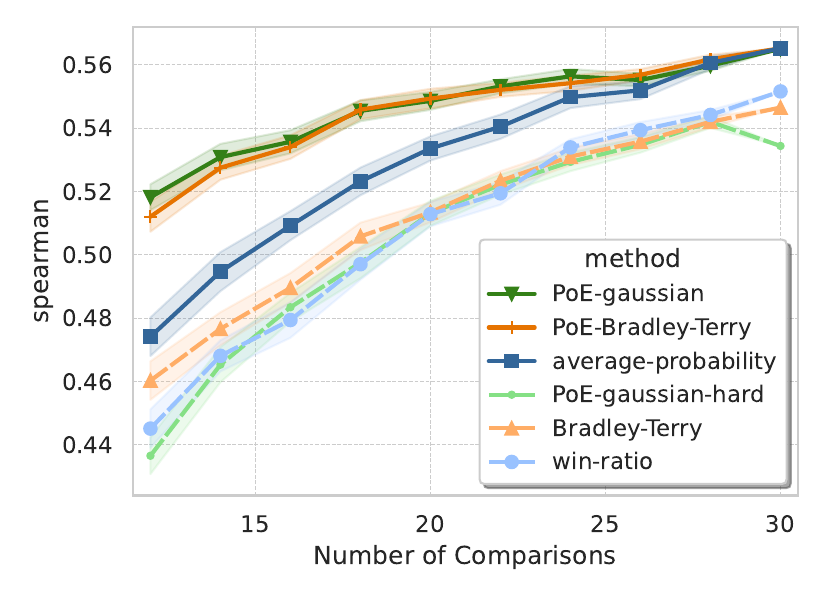}
        \caption{\smallcaption FlanT5-11B, TopicalChat \texttt{COH}}
    \end{subfigure}%
    ~ 
    \begin{subfigure}[t]{0.33\textwidth}
        \centering
         \includegraphics[width=\textwidth]{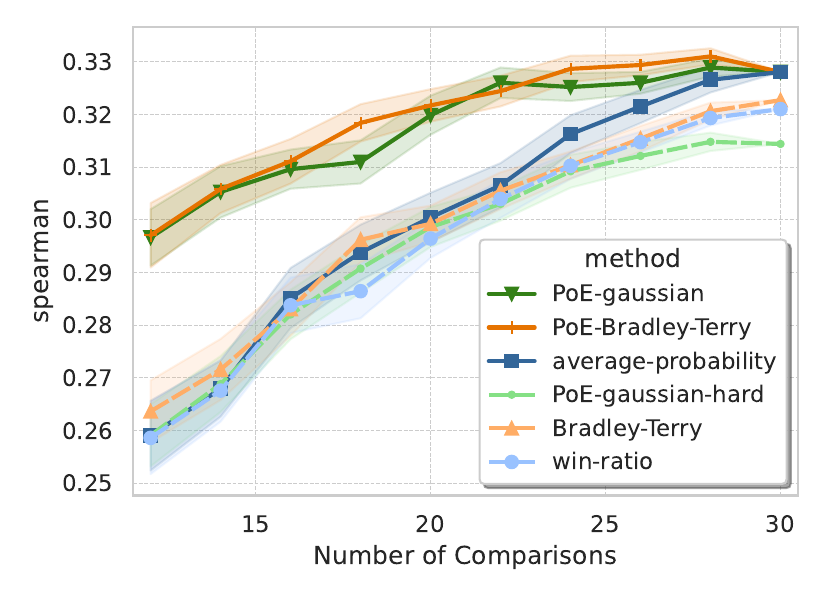}
        \caption{\smallcaption Llama-7B, TopicalChat \texttt{CNT}}
    \end{subfigure}%
    ~ 
    \begin{subfigure}[t]{0.33\textwidth}
        \centering
         \includegraphics[width=\textwidth]{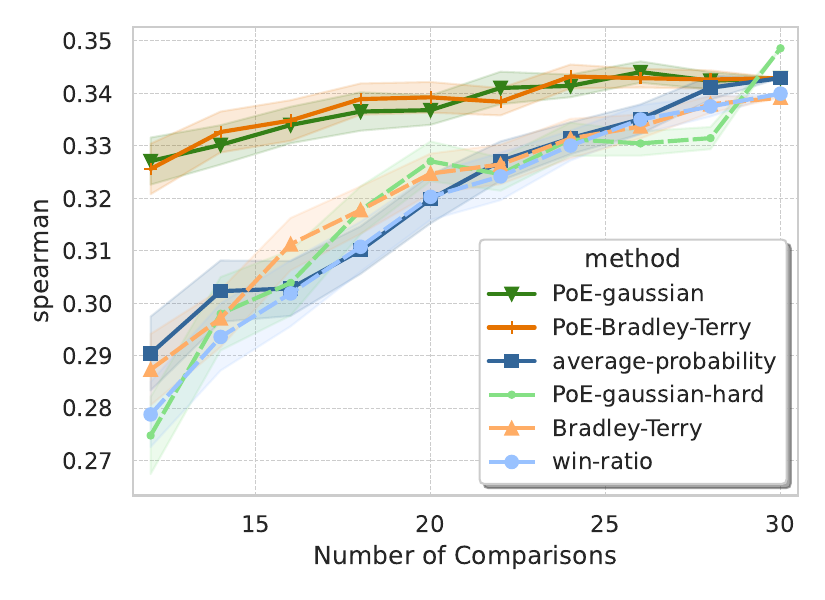}
        \caption{\smallcaption Llama2-13B, TopicalChat \texttt{NAT}}
    \end{subfigure}
    \\
    \caption{Efficiency curves when sweeping $K$, the number of comparisons per context, where at each $K$ the comparisons are randomly drawn 100 times. Average performance with 95\% confidence is displayed. These curves were randomly selected from all possible configurations.} 
    \label{fig:appendix_summeval_topicalchat}
\end{figure}

Figure \ref{fig:appendix_summeval_topicalchat} showcases the performance of the various scoring approaches for further models/attributes for SummEval and TopicalChat. We observe that in all cases the PoE approaches lead to best performance when only a subset of comparisons are used.

\vspace{-2mm}

\newpage
\subsection{HANNA and CMCQRD Chat Efficiency Plots}
\label{ssec:extra_efficient_plots_hanna_cmcqrd}

\begin{figure}[h]
    \centering
    \begin{subfigure}[t]{0.33\textwidth}
         \centering
         \includegraphics[width=\textwidth]{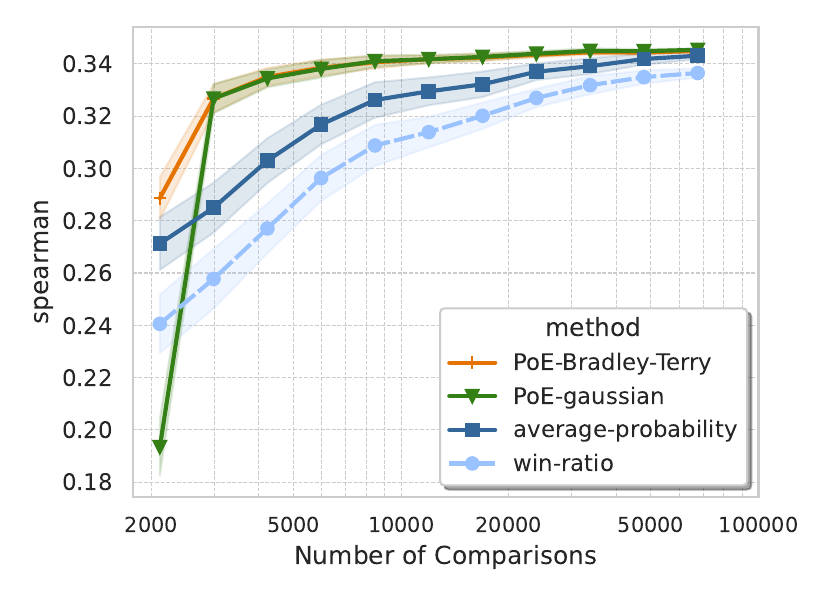}
        \caption{\smallcaption Llama2-7B, HANNA \texttt{SUR}}
    \end{subfigure}%
    ~ 
    \begin{subfigure}[t]{0.33\textwidth}
        \centering
         \includegraphics[width=\textwidth]{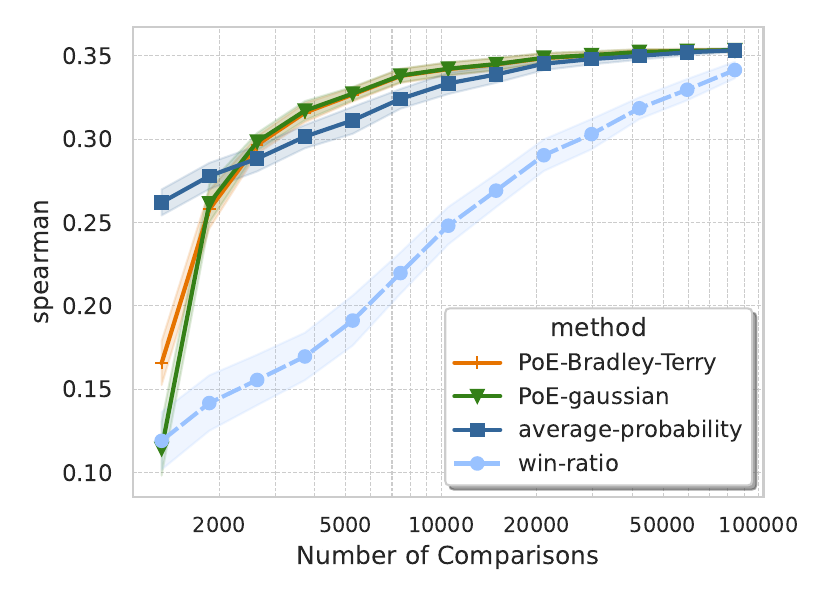}
        \caption{\smallcaption Llama2-7B, CMCQRD \texttt{DIF}}
    \end{subfigure}%
    ~ 
    \begin{subfigure}[t]{0.33\textwidth}
        \centering
         \includegraphics[width=\textwidth]{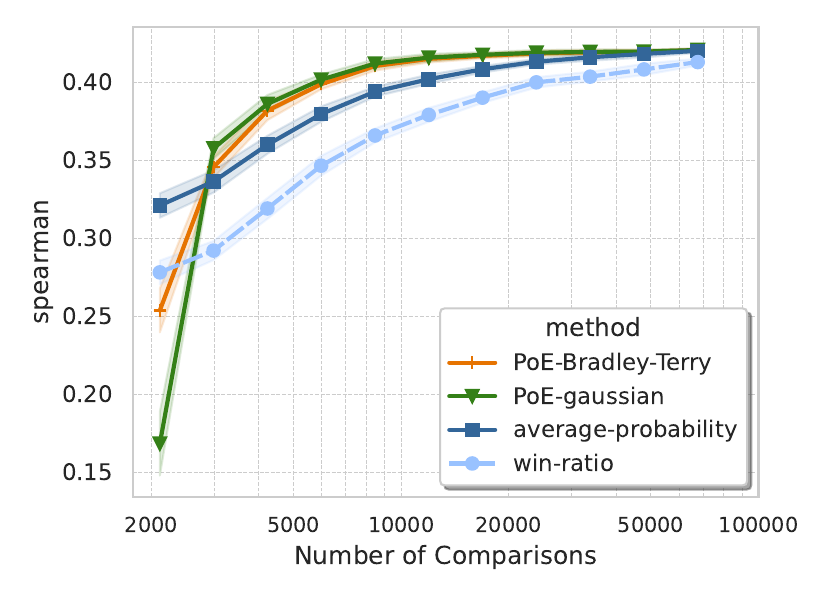}
        \caption{\smallcaption Llama-13B, HANNA \texttt{COH}}
    \end{subfigure} \\
    \begin{subfigure}[t]{0.33\textwidth}
         \centering
         \includegraphics[width=\textwidth]{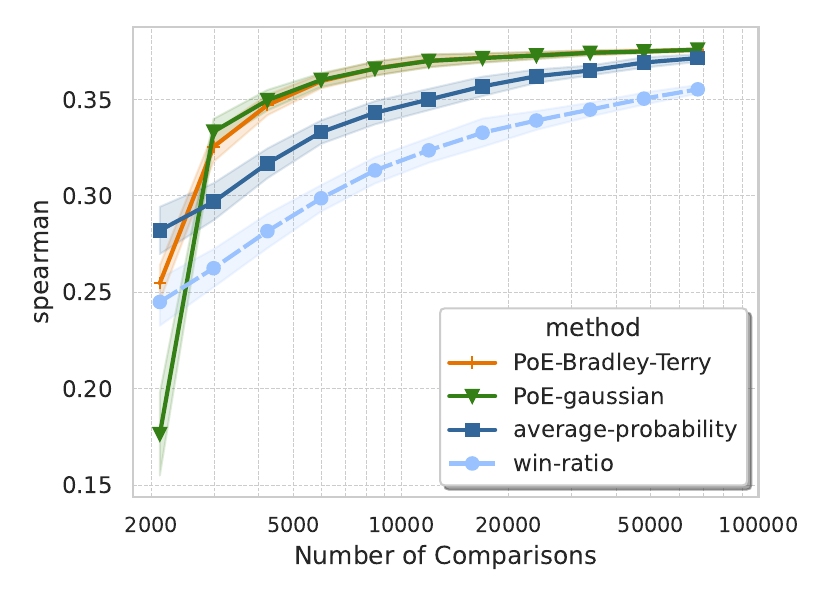}
        \caption{\smallcaption Llama-13B, HANNA \texttt{SUR}}
    \end{subfigure}%
    ~ 
    \begin{subfigure}[t]{0.33\textwidth}
        \centering
         \includegraphics[width=\textwidth]{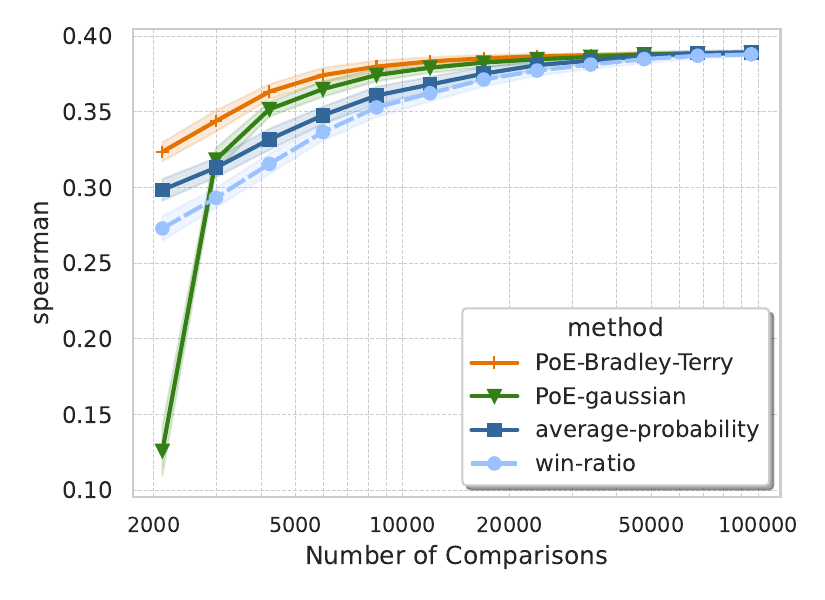}
        \caption{\smallcaption Mistral-7B, HANNA \texttt{COH}}
    \end{subfigure}%
    ~ 
    \begin{subfigure}[t]{0.33\textwidth}
        \centering
         \includegraphics[width=\textwidth]{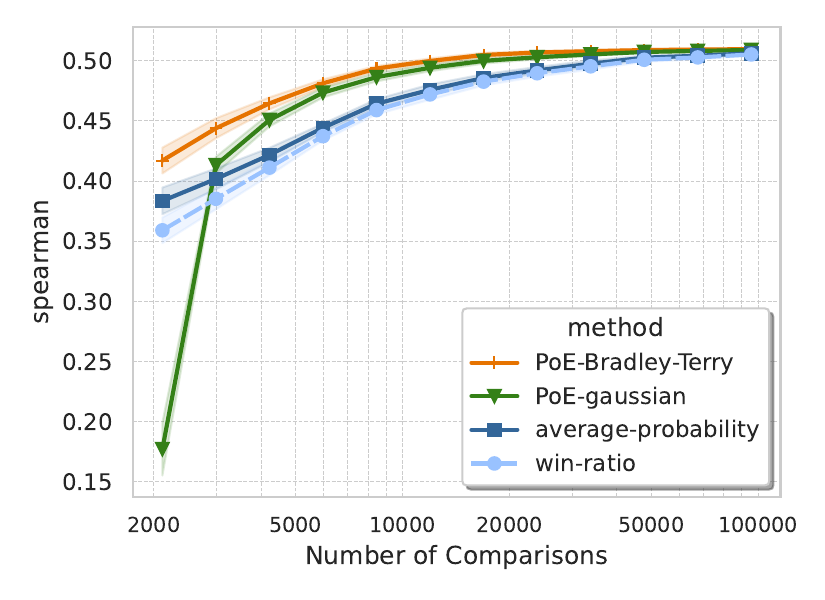}
        \caption{\smallcaption Mistral-7B, HANNA \texttt{CMP}}
    \end{subfigure}
    \\
    \caption{Efficiency curves where comparisons are randomly drawn 20 times. These curves were randomly selected from all possible configurations.} 
    \label{fig:appendix_hanna_cmcqrd_curves}
\end{figure}

Figure \ref{fig:appendix_hanna_cmcqrd_curves} showcases further performance curves for HANNA and CMCQRD, which demonstrate the effectiveness of the PoE framework in further settings with large $N$.

\subsection{Non-Symmetric Efficiency Plots}
\begin{figure}[h]
    \centering
    \begin{subfigure}[t]{0.4\textwidth}
         \centering
         \includegraphics[width=\textwidth]{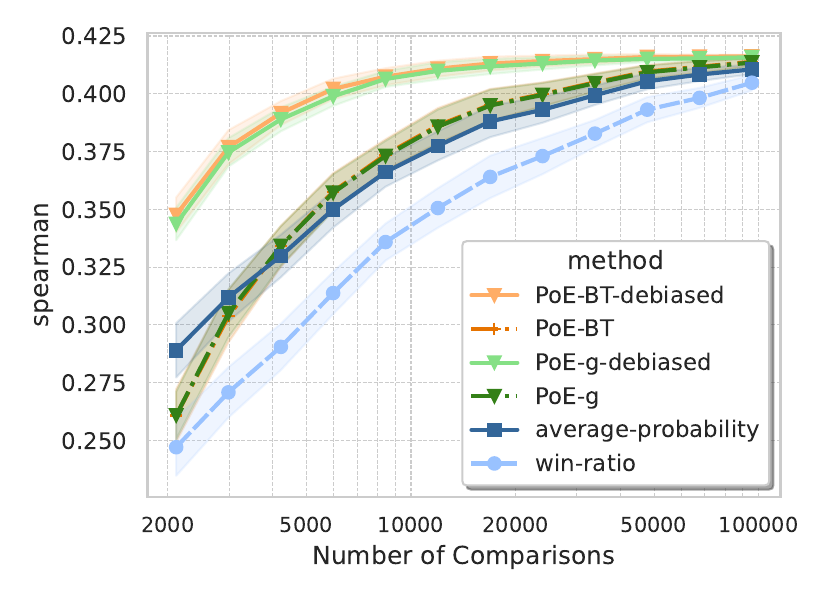}
        \captionsetup{format=plain, justification=centering}
        \caption{Llama-7B, HANNA \texttt{COH}, \\non-symmetric}
    \end{subfigure}%
    ~ 
    \begin{subfigure}[t]{0.4\textwidth}
        \centering
         \includegraphics[width=\textwidth]{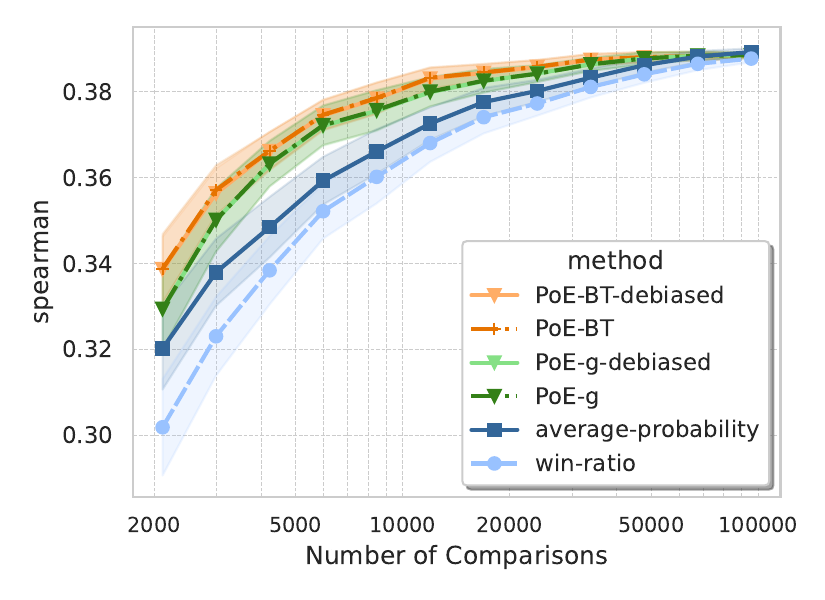}
        \captionsetup{format=plain, justification=centering}
        \caption{Mistral-7B, HANNA \texttt{COH}, \\ symmetric vs non-symmetric}
    \end{subfigure} \\
    \caption{Efficiency curves in the non-symmetric set-up.} 
    \label{fig:appendix_unbalanced}
\end{figure}

Figure \ref{fig:appendix_unbalanced} shows the performance curves for Llama-7B and Mistral 7B. Mistral-7B has minimal positional bias with $E[p_{ij}]\!=\!0.51$, while Llama-7B has considerable bias with $E[p_{ij}]\!=\!0.78$. For Llama2-7B, the debiased experts, ${\tt p}_{\gamma}(s_i - s_j | p_{ij})$, yield large performance gains and performance does not converge quickly without it. For Mistral-7B, the debiasing parameter has little influence, as expected since $\gamma$ will be near 0. Note that, although Llama2-7B is more biased, it has better judgement capabilities and achieves better correlations, though the debiasing parameter is required.

\subsection{Symmetric vs Non-Symmetric Efficiency Plots}
\label{ssec:app_sym_vs_nonsym}

\begin{figure}[h]
    \centering
    \begin{subfigure}[t]{0.33\textwidth}
         \centering
         \includegraphics[width=\textwidth]{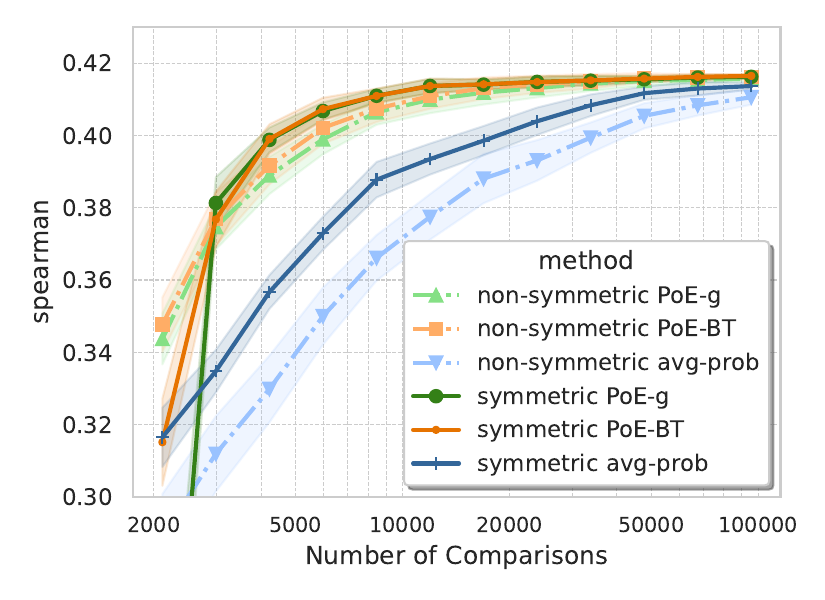}
        \captionsetup{format=plain, justification=centering}
        \caption{\smallcaption Llama2-7B, HANNA \texttt{COH}, \\ symmetric vs non-symmetric}
    \end{subfigure}%
    ~ 
    \begin{subfigure}[t]{0.33\textwidth}
        \centering
         \includegraphics[width=\textwidth]{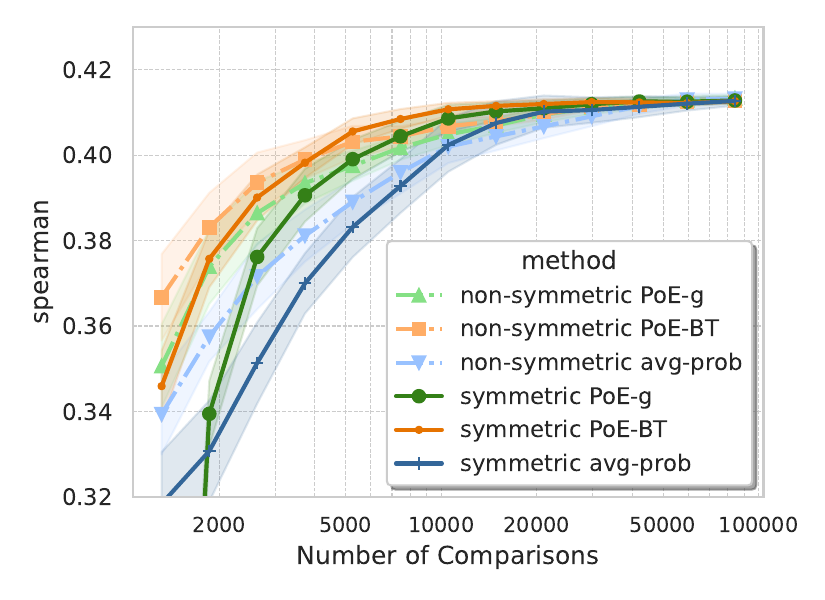}
        \captionsetup{format=plain, justification=centering}
        \caption{\smallcaption Mistral-7B, CMCQRD \texttt{DIF}, \\ symmetric vs non-symmetric}
    \end{subfigure}%
    ~ 
    \begin{subfigure}[t]{0.33\textwidth}
        \centering
         \includegraphics[width=\textwidth]{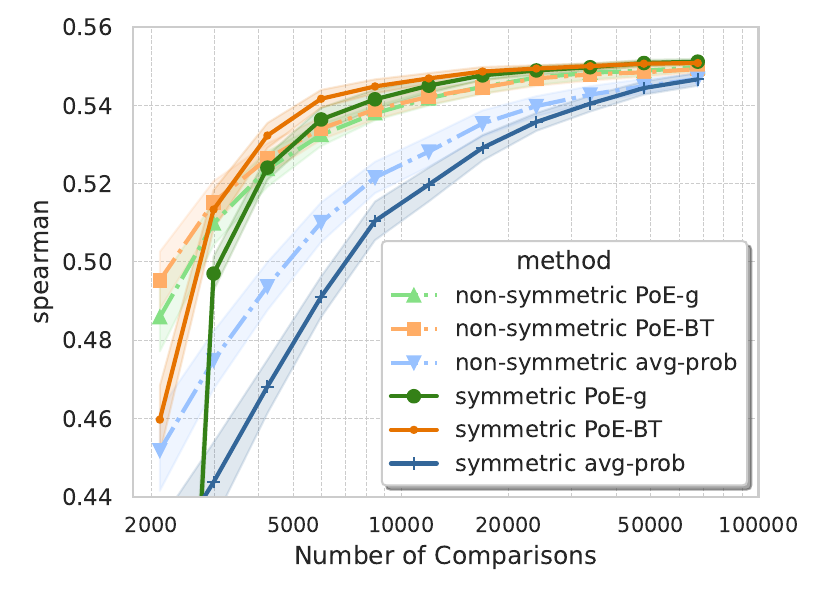}
        \captionsetup{format=plain, justification=centering}
        \caption{\smallcaption Llama2-13B, HANNA \texttt{CMP}, \\ symmetric vs non-symmetric}
    \end{subfigure} \\
    \caption{Efficiency Curves when sweeping $K$, the number of comparisons per context, with 95\% confidence intervals using 100 samples per step for non-symmetric set-up. These curves were randomly selected from all possible configurations.} 
    \label{fig:appendix_unbalanced_vs_balanced}
\end{figure}

For several other models and datasets, Figure \ref{fig:appendix_unbalanced_vs_balanced} compares the performance between symmetric and non-symmetric attributes, as well as against the average probability and win-ratio. We observe that both perform well and often similarly, although minor differences in characteristics can be observed, as discussed in the main paper.

\subsection{Data Analysis}
\begin{figure}[h]
    \centering
    \begin{subfigure}[t]{0.4\textwidth}
        \centering
         \includegraphics[width=\textwidth]{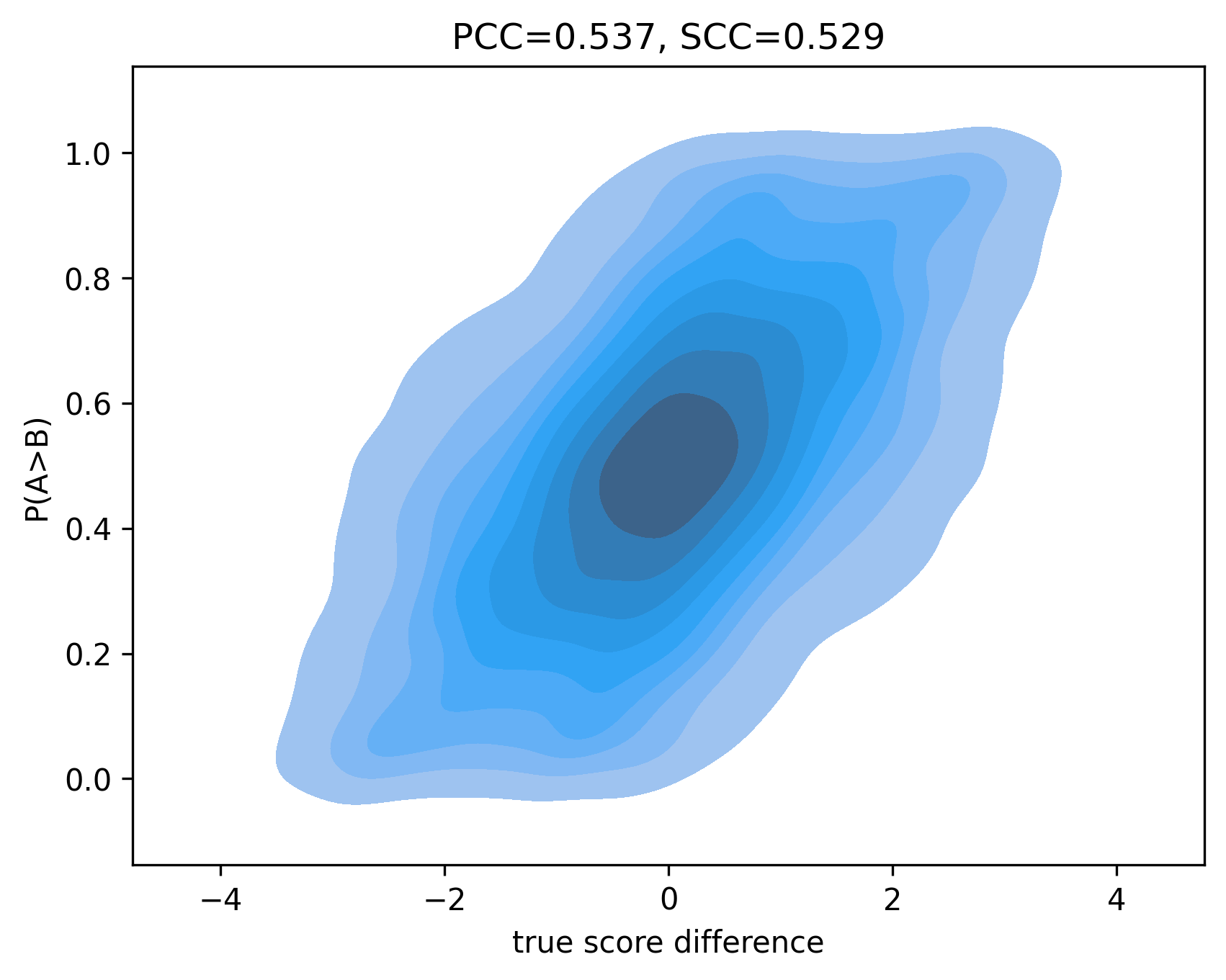}
        \caption{\smallcaption FlanT5-3B, SummEval \texttt{COH}}
    \end{subfigure}%
    ~ 
    \begin{subfigure}[t]{0.4\textwidth}
        \centering
         \includegraphics[width=\textwidth]{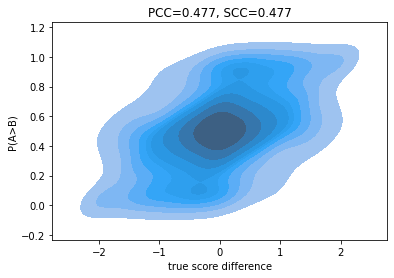}
        \caption{\smallcaption Llama2-13B, TopicalChat \texttt{CNT}}
    \end{subfigure}%
    \\
    \caption{Joint distribution of the LLM probabilities and true scores.}
    \label{fig:bivariate}
\end{figure}

In the PoE framework, each expert models the distribution ${\tt p}(s_i \!-\! s_j | p_{ij})$. To determine a suitable form of the expert, and whether the Gaussian and/or the extended Bradley-Terry experts are sensible assumptions, Figure \ref{fig:bivariate} displays the joint bivariate distribution between the true score difference $s_i \!-\! s_j$ and the observed probability $p_{ij}$. For a particular LLM, all comparisons over all the contexts of the dataset are assessed. The frequency count of the LLM probability and true score difference (calculated using the gold-standard annotator labels) is then plotted. The plots illustrate a clear correlation between the probabilities and score difference, implying that considerable scoring information can be gained from leveraging probabilities and decisions. However, the mapping is not deterministic, and there is considerable noise present. Empirically, The distributions appear to be well approximated by Gaussian distributions, implying that the conditional distributions will also be well-modelled by Gaussian distributions. 

\begin{figure}[h]
    \centering
    \begin{subfigure}[t]{0.4\textwidth}
        \centering
         \includegraphics[width=\textwidth]{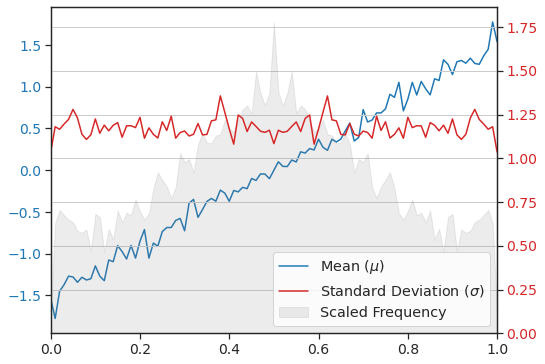}
        \caption{\smallcaption FlanT5-3B, SummEval \texttt{COH}}
    \end{subfigure}%
    ~ 
    \begin{subfigure}[t]{0.4\textwidth}
        \centering
         \includegraphics[width=\textwidth]{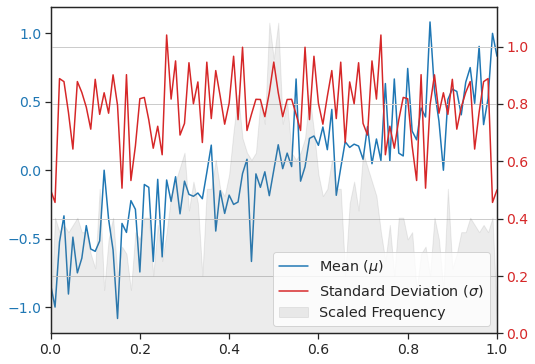}
        \caption{\smallcaption Llama2-13B, TopicalChat \texttt{CNT}}
    \end{subfigure}%
    \\
    \caption{Expected score difference and variance given the LLM probability.} \label{fig:mean_std}
    \label{fig:lienar_assumptions}
\end{figure}

We further analyze the relationship between the LLM probability $p$ and the expected score difference, $\delta(p) = E_{p_{ij}}[s_i \!-\! s_j \;|\; |p_{ij}  \!-\! p| \!<\! \epsilon]$. Figure \ref{fig:mean_std} demonstrates that 1) the probability is quite linearly correlated with the expected score difference; and 2) the variance across all score distributions given the probability is quite constant. Therefore the Gaussian assumptions discussed in Section \ref{ssec:gaussian_assumptions} appear to be reasonable. Note that TopicalChat is a smaller dataset (with 1800 total comparisons) and hence has more observed noise.

\subsection{Comparison Against Additional baselines}
Throughout the paper, baselines such as the Bradley Terry, average probability and win-ratio were used as methods to compare the best method to get scores from comparative outcomes. However alternate methods are possible, which do not necessarily combine information from a subset of the comparisons. For example, G-EVAL \cite{liu2023g} uses a prompt that asks the model to directly score texts and then calculates the fair mean over the probabilities of scores. While PairS \cite{liu2024aligning} considers sorting algorithms to guide which pairwise comparisons should be made, as well as for determining the final rankings. Table \ref{tab:pairs_geval_baselines} displays the performance of our Product of Experts Framework of LLM comparative assessment against these baselines for SummEval and HANNA (using a modest $K=3N$ and $K=5N$ respectively) and demonstrates that our approach has considerably better performance over the other baseline methods, where in 11/14 settings has the best performance (and often by considerable margins). 
\begin{table}[h]
    \centering
    \small
    \renewcommand\tabcolsep{4pt}
    \begin{tabular}{cl|cccc|ccc}
    \toprule
    & & \multicolumn{4}{|c|}{SummEval} & \multicolumn{3}{c}{HANNA} \\
    & K & COH & CON & FLU & REL & COH & CMP & SUR \\ 
    \midrule
    \multirow{4}{*}{Llama2-7B}        
    & G-Eval        & 15      & 23      & 7       & 20      & 25      & 33      & 17\\
    %& PAIRS-greedy  & 16      & 28      & 16      & 20      & -       & -       & -\\
    & PAIRS-beam    & 17      & \bf{31} & 18      & 24      & 29      & 17      & 19\\
    & PoE-BT        & \bf{29} & 24      & \bf{20} & \bf{34} & \bf{41} & \bf{48} & \bf{34}\\
    \midrule
    \multirow{4}{*}{Mistral-7B}       
    & G-Eval       & 25       & \bf{39} & 20      & 25      & 34      & 39      & 25 \\
    %& PAIRS-greedy & 25       & 25      & 8       & 29      & -       & -       & - \\
    & PAIRS-beam   & 28       & 30      & 24      & 27      & 33      & 31      & \bf{27}\\
    & PoE-BT       & \bf{34} & 36      & \bf{26} & \bf{37} & \bf{38} & \bf{50} & 26\\
    \bottomrule
    \end{tabular}
    \vspace{1mm}
    \caption{SummEval performance for SummEval and HANNA for all particular attributes.}
    \label{tab:pairs_geval_baselines}
    \vspace{-0.5cm}
\end{table}

\end{document}